\documentclass{article}

\usepackage[final]{neurips_2024}

\usepackage{graphicx}
\usepackage{amsmath, amsthm, amssymb}

\usepackage[utf8]{inputenc} % allow utf-8 input
\usepackage[T1]{fontenc}    % use 8-bit T1 fonts
\usepackage{hyperref}       % hyperlinks
\usepackage{url}            % simple URL typesetting
\usepackage{booktabs}       % professional-quality tables
\usepackage{amsfonts}       % blackboard math symbols
\usepackage{nicefrac}       % compact symbols for 1/2, etc.
\usepackage{microtype}      % microtypography
\usepackage[dvipsnames]{xcolor}         % colors

\usepackage{caption}
\usepackage{subcaption}

\usepackage{algorithm, algorithmic}
\usepackage{cleveref}

\usepackage{makecell}

\usepackage{wrapfig}

\usepackage[utf8]{inputenc}
\usepackage{ucs}
\usepackage[T1]{fontenc}
\usepackage{textcomp}

\newcommand{\Dprior}{D_{\text{prior}}}
\newcommand{\Ddown}{D_{\text{down}}}

\newcommand{\Dtrain}{D_{\text{train}}}

\newcommand{\thetaprior}{\theta_{\text{prior}}}
\newcommand{\thetapriordown}{\theta_{\text{prior}+ \text{down}}}
\newcommand{\thetapriordownx}{\theta_{\text{prior}+ \text{down}+x_{<i}}}
\newcommand{\thetapriorx}{\theta_{\text{prior}+x_{<i}}}
\newcommand{\thetadown}{\theta_{\text{down}}}
\newcommand{\thetadownx}{\theta_{\text{down}+x_{<i}}}
\newcommand{\thetax}{\theta_{x_{<i}}}

\newcommand{\thetacond}{\theta^{\text{cond}}}
\newcommand{\thetamarg}{\theta^{\text{marg}}}

\title{
{\color{Red}C}{\color{Orange}o}{\color{Goldenrod}L}{\color{Green}o}{\color{NavyBlue}R}{\color{NavyBlue}-}{\color{Purple}Filter}:
Conditional Loss Reduction Filtering \\for Targeted Language Model Pre-training
}

\author{
David Brandfonbrener\\
Kempner Institute at Harvard University
\And
Hanlin Zhang\\
Harvard University
\And
Andreas Kirsch\\
University of Oxford
\And 
Jonathan Richard Schwarz\\
Harvard University
\And
Sham Kakade\\
Kempner Institute at Harvard University
}

\definecolor{caribbeangreen}{rgb}{0.0, 0.8, 0.6}

\begin{document}

\maketitle

\begin{abstract}
Selecting high-quality data for pre-training is crucial in shaping the downstream task performance of language models. A major challenge lies in identifying this optimal subset, a problem generally considered intractable, thus necessitating scalable and effective heuristics. In this work, we propose a data selection method, CoLoR-Filter (Conditional Loss Reduction Filtering), which leverages an empirical Bayes-inspired approach to derive a simple and computationally efficient selection criterion based on the relative loss values of two auxiliary models.

In addition to the modeling rationale, we evaluate CoLoR-Filter empirically on two language modeling tasks: (1) selecting data from C4 for domain adaptation to evaluation on Books and (2) selecting data from C4 for a suite of downstream multiple-choice question answering tasks. We demonstrate favorable scaling both as we subselect more aggressively and using small auxiliary models to select data for large target models. As one headline result, CoLoR-Filter data selected using a pair of 150m parameter auxiliary models can train a 1.2b parameter target model to match a 1.2b parameter model trained on 25b randomly selected tokens with 25x less data for Books and 11x less data for the downstream tasks. 

Code: {\tiny \url{https://github.com/davidbrandfonbrener/color-filter-olmo}}

Filtered data: {\tiny \url{https://huggingface.co/datasets/davidbrandfonbrener/color-filtered-c4} }
\end{abstract}

\section{Introduction}

\begin{figure}[t]
    \centering
    \includegraphics[width=\textwidth]{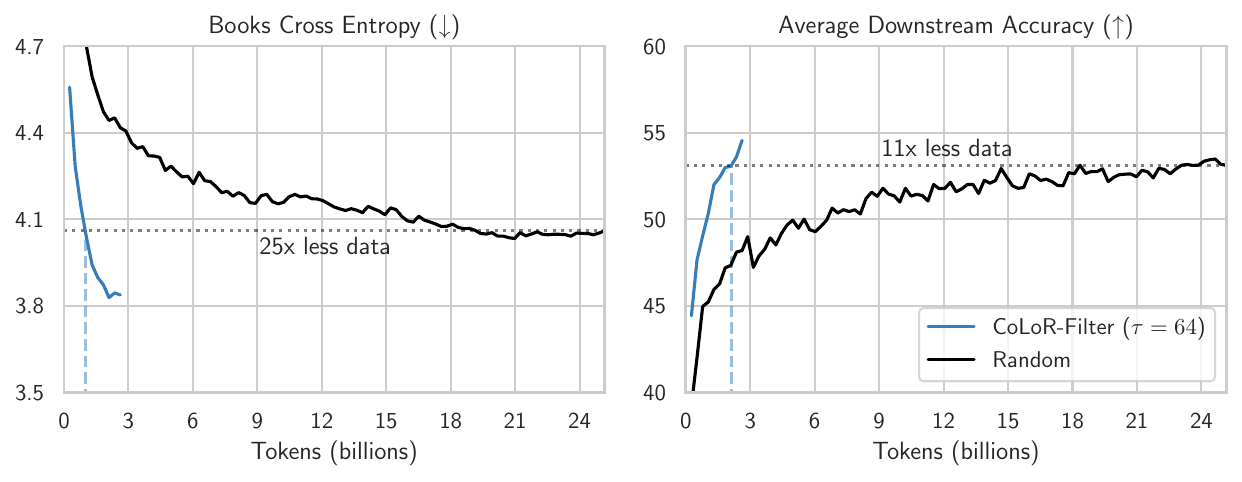}
    \vspace{-0.2cm}
    \caption{Learning curves for 1.2 billion parameter language models trained on data selected by CoLoR-Filter using smaller 150 million parameter auxiliary models for two different target distributions. (Left) We target and evaluate loss on Books, lower is better. (Right) We target and evaluate accuracy on a suite of 8 downstream tasks from \citep{groeneveld2024olmo}, higher is better. In both cases, test data is held out from the data used by CoLoR-Filter to guide selection. $\tau$ is the subset size multiplier denoting the number of examples considered for each selected data point. The CoLoR-Filter line terminates when we run out of data in C4 ($\approx$175b possible tokens).}
    \label{fig:1b}
\end{figure}

The content of the data that a language model is trained on can have profound effects on its performance and the efficiency of the training process \citep{rae2021scaling, longpre2023pretrainer, penedo2023refinedweb, cerebras2023slimpajama, li2024datacomp}. But it remains an open research question how to decide which data to include in the training set.
In this paper, we analyze a family of loss-based approaches for targeted selection of pre-training data, propose a simple approach that outperforms existing methods, and provide some preliminary evidence of favorable scaling properties. 

%When pre-training a language model, there is always some budget of training compute. This budget constrains the total number of sequences that one can use as training data. In this paper we approach this issue from first principles and propose a simple algorithm for selecting training sequences.

To formulate the data selection problem, we first need to specify an objective that quantifies whether the selected data is good. 
Defining this objective requires evaluating a pre-trained language model, which is an area of active research \citep{eval-harness, magnusson2023paloma, engstrom2024dsdm, chang2024survey}. 
For this paper, we will take the goal to be to maximize performance on a set of downstream tasks.
Since the preferred metrics on a given set of tasks are not necessarily the same nor amenable to direct optimization, we consider the likelihood of sequences sampled from the downstream tasks as a proxy objective.  
With this objective, we now have a straightforward goal: 
given a very large corpus of sequences and a small amount of high-quality data from a set of downstream tasks, we want to select a subset from the corpus so that training on the selected data maximizes likelihood on the downstream tasks. 
Then we can also test performance on the tasks under their preferred metrics.

From this objective, we derive an algorithm dubbed CoLoR-Filter (Conditional Loss Reduction Filtering). In \Cref{sec:derivations} we derive this method by applying Bayes' rule and approximate empirical Bayes to the downstream likelihood objective. The resulting method is simple and intuitive: each sequence is scored by the difference in likelihood between a ``prior'' model and a ``conditional'' model that results from fine-tuning the prior model on the downstream data. Sequences that are more likely under the fine-tuned model are good. We also compare this algorithm to prior work (e.g., \citep{mindermann2022prioritized}) and discuss computational costs.

To evaluate our method, we consider two tasks. First, in \Cref{sec:books}, we consider a semi-synthetic task where the downstream task is language modeling on Books. Given access to C4 \citep{raffel2020exploring} as potential pre-training data and a small (25 million tokens) sample of data from Books, we use CoLoR-Filter and a variety of baselines to select 3 billion tokens. We find that data selected by CoLoR-Filter can substantially outperform models trained on 8x as much randomly chosen data.
Second, in \Cref{sec:down}, we consider a suite of 8 downstream multiple-choice tasks from \citet{groeneveld2024olmo}. As downstream data we take the training sets of the tasks, but we evaluate accuracy on the held-out test sets. We again find that selecting with CoLoR-Filter outperforms training on 8x as much randomly selected data. Moreover, in both tasks, performance scales smoothly with the hyperparameter $ \tau $ that governs how aggressively we select the data, suggesting that further scaling would yield further improvements.

In addition to finding that CoLoR-Filter can select good subsets of data, we also consider the computational cost of the selection procedure itself. CoLoR-Filter only requires running inference of the two auxiliary models to select data. This is computationally beneficial compared to online methods like RHOLoss \citep{mindermann2022prioritized} since inference is cheaper than training and is entirely parallelizable. To maximize the computational benefits we also show that data selected with a small (150 million parameter) model can be transferred to a larger (1.2 billion parameter) model. Results are shown in \Cref{fig:1b}, showing substantial efficiency improvements. 

%Taken together, our results demonstrate the feasibility of scalable data selection for language modeling. We show that lightly curated web-scraped datasets like C4 contain small subsets that can yield better performance on particular downstream tasks and that with CoLoR-Filter we can find these subsets efficiently.

\section{Setting and Derivations}\label{sec:derivations}

Assume that we are given a large pre-training dataset $ \Dtrain$, a small downstream dataset $ \Ddown$ from the downstream task(s) of interest, and a ``prior'' dataset $ \Dprior$ we can use as prior knowledge (in practice we often just sample from $ \Dtrain$). 
We will assume for all practical purposes that $ \Dtrain$ is infinite and training proceeds in the ``online'' or ``single pass'' setting where we do not repeat data points.
Our goal is to choose a subset $ S \subset \Dtrain$ of a fixed size $ |S| = n $ that minimizes the downstream loss (maximizes the downstream likelihood).

This section introduces our CoLoR-Filter algorithm, inspired by and building upon the RHOLoss approach from prior work ~\citep{mindermann2022prioritized,evans2023bad}. We also discuss related algorithms applicable to this setting such as DSIR \citep{xie2023data} and DSDM \citep{engstrom2024dsdm}. Additional related work is discussed further in \Cref{sec:related}.

\subsection{Bayesian Data Selection}

Our objective can be formulated as a Bayesian optimization problem, where the goal is to select a set $S$ so as to maximize the posterior probability of $\Ddown$, i.e.
\begin{align}
    \min_{S \subset \Dtrain, |S| = n} -\log \Pr (\Ddown | S),
\end{align}
where $\Pr (\Ddown | S)$ is the posterior probability.
Applying Bayes rule we get:
\begin{align}
    \min_{S \subset \Dtrain, |S| = n} -\log \Pr(S| \Ddown) + \log \Pr(S) - \log \Pr(\Ddown)
\end{align}
Note that the last term does not depend on $ S$, so it can be ignored when optimizing over $ S$. Introducing a prior over model parameters $ \theta$, we get:
\begin{align}\label{eq:bayes}
     \min_{S \subset \Dtrain, |S| = n} \underbrace{-\log \int_\theta\Pr(S| \theta)\Pr(\theta | \Ddown)}_{\text{``conditional''}} + \underbrace{\log \int_\theta \Pr(S | \theta)\Pr(\theta)}_{\text{``marginal''}}
\end{align}
We will refer to the two terms as the conditional and marginal terms, respectively.\footnote{Prior work \citep{mindermann2022prioritized, evans2023bad} has referred to the models that estimate these two terms as the ``reference'' and ``learner'' or ``actor'', respectively. We opt for the names conditional and marginal for clarity in connections to the Bayesian viewpoint.} Note that the conditional and marginal terms together make up the negative pointwise mutual information between the selected and downstream data, which has deep connections to prior work on active learning and active sampling \citep{lindley1956measure, moore2010intelligent, houlsby2011bayesian, bickfordsmith2023prediction, kirsch2023a, rainforth2024modern}.

\subsection{CoLoR-Filter}\label{sec:color-filter}
Given that we have access to prior knowledge from the dataset $ \Dprior$, we can replace the uninformed prior over $ \theta $ with an empirical Bayes prior that conditions on $\Dprior$ to obtain:
\begin{align}
     \min_{S \subset \Dtrain, |S| = n} -\log \int_\theta\Pr(S| \theta)\Pr(\theta | \Ddown, \Dprior) + \log \int_\theta \Pr(S | \theta)\Pr(\theta| \Dprior)
\end{align}
As this integration is still intractable, we now make our main simplifying assumption which is to replace this integration over parameters by a point estimate: 
\begin{align}\label{eq:prior}
     \approx \min_{S \subset \Dtrain, |S| = n}  -\log \Pr(S | \thetapriordown) + \log \Pr(S | \thetaprior),
\end{align}
where $ \thetaprior$ is a model trained on $ \Dprior$ and $ \thetapriordown$ is a model trained on both $ \Dprior$ and  $\Ddown$ (in practice, we use a model that is pre-trained on $ \Dprior$ fine-tuned on $ \Ddown$). 

Moreover, this approximation leads to computational benefits by avoiding the full combinatorial optimization of subset selection. In particular, once we condition on a single model $ \theta$, and assuming the distribution over points $ x \in S$ is independent, i.e. $ \Pr(S|\theta) = \prod_{x \in S} \Pr(x|\theta)$, we have:
\begin{align}
    \min_{\{x_1,\dots, x_{n}\} \subset \Dtrain}   -\log \prod_{i=1}^n \Pr(x_i| \thetapriordown) + \log \prod_{i=1}^n \Pr(x_i | \thetaprior)
\end{align}
which simplifies to:
\begin{align}
     \min_{\{x_1,\dots, x_{n}\} \subset \Dtrain}  \sum_{i=1}^n -\log \Pr(x_i| \thetapriordown) -  (-\log \Pr(x_i | \thetaprior))
\end{align}
This gives our CoLoR-Filter criteria that we use to select data. This optimization selects the points with the largest conditional loss reduction (CoLoR), i.e. the points where the negative log-likelihood loss of the conditional model $ \thetapriordown$ is lower than the marginal model $ \thetaprior$.
Intuitively, this selects data points that are more likely under the conditional model than the marginal model.

\paragraph{A note on data diversity.}
While the factorization that results from our point estimate of the parameters is computationally convenient, it makes an important simplifying assumption. In particular, the CoLoR-Filter objective no longer encourages the selection of a diverse dataset, as scores are applied independently to each point. In practice, this is remedied by a few considerations: (1) we can run CoLoR-Filter on a corpus that has already been deduplicated to prevent degenerate duplications, (2) for large $ n$, we must select many different data points, and (3) each datapoint is itself a sequence that may contain diverse signal across tokens. We should also note this is not a unique property of CoLoR-Filter and also happens in other methods that do offline scoring like DSDM and DSIR.
We defer a detailed discussion of the nuances of this issue to \Cref{app:online}.

\subsection{Related Algorithms}\label{sec:related_algs}

\paragraph{Connection to importance sampling.} 
Since the CoLoR-Filter objective is written as a difference of logs, it can also be written as a log of the ratio between probabilities under $ \thetapriordown$ and $ \thetaprior$.
If data were actually sampled from $ \thetaprior$, then this ratio would be the importance weight needed to reweight samples so that they are from the model defined by $ \thetapriordown$.
Note that DSIR \citep{xie2023data} directly attempts to perform importance sampling from $ \Dtrain$ to $ \Ddown$ instead of optimizing performance on the downstream data. Thus, DSIR ends up with a somewhat related algorithm except in DSIR: (1) there is no language model, just features of a full data point (hashed n-grams), and (2) the algorithm samples rather than optimizes.

\paragraph{Connections to DSDM.} 
Another closely related approach is DSDM \citep{engstrom2024dsdm} which uses a TRAK Datamodel estimator \citep{ilyas2022datamodels, park2023trak} to score datapoints and then selects the top-$n$ points. The motivation and setting of DSDM are similar to CoLoR-Filter, but DSDM relies on TRAK which constructs a linear approximation of the influence that data points have on each other. Instead, CoLoR-Filter operates directly in function space by comparing the loss between models directly rather than relying on linear approximations or Datamodels \citep{ilyas2022datamodels}.

%\subsection{RHO-down}\label{sec:rho}

\paragraph{Connections to RHO-down.} 
CoLoR-Filter is inspired by and builds on the RHOLoss approach introduced in prior work \citep{mindermann2022prioritized} with subtle but significant differences in the setting: the original RHO paper focuses on cases where the hold-out data is sampled from the same distribution as $ \Dtrain$ over multiple epochs of training. In contrast, we focus on selecting data to target downstream distributions that are different from $ \Dtrain$ and where we only take a single pass over the data. Here, we derive a straightforward adaptation of RHOLoss to our setting, which we call RHO-down.

We now derive RHO-down in our setting, aiming to illustrate the connections between RHO-down and CoLoR-Filter.
First, RHO-down approximates the full subset selection problem from \Cref{eq:bayes} by a greedy (sequential) approximation where samples are added to $ S $ one (batch) at a time. Using a batch size of $1$, the $i$th-sample would be ideally added according to the following criterion:
\begin{align}
     \approx \min_{x_i \in \Dtrain}   -\log \int_\theta\Pr(x_i| \theta)\Pr(\theta | \Ddown, x_{<i}) + \log \int_\theta \Pr(x_i | \theta)\Pr(\theta| x_{<i}),
\end{align}
where $i$ ranges from $1$ to $n$ sequentially.
RHO-down then uses a point estimate of the parameters (as we do in CoLoR-Filter):
\begin{align}
     \approx \min_{x_i \in \Dtrain}  -\log \Pr(x_i| \thetadownx) + \log  \Pr(x_i | \thetax)
\end{align}
Finally, the RHO-down authors found that updating the conditional term to depend on $ x_{<i}$ was unstable, so they instead approximate this by a fixed model $ \thetadown$:
\begin{align}
     \approx \min_{x_i \in \Dtrain} -\log \Pr(x_i| \thetadown) + \log  \Pr(x_i | \thetax).
\end{align}

\iffalse
First, RHO-down approximates the full subset selection problem from \Cref{eq:bayes} by a greedy approximation where samples are added to $ S $ one (batch) at a time. 
\begin{align}
     \approx \min_{x_1, \dots, x_n \subset \Dtrain} \sum_{i=1}^n  -\log \int_\theta\Pr(x_i| \theta)\Pr(\theta | \Ddown, x_{<i}) + \log \int_\theta \Pr(x_i | \theta)\Pr(\theta| x_{<i})
\end{align}
From here, RHO-down uses a point estimate of the parameters (as we do in CoLoR-Filter):
\begin{align}
     \approx \min_{x_1, \dots, x_n \subset \Dtrain} \sum_{i=1}^n  -\log \Pr(x_i| \thetadownx) + \log  \Pr(x_i | \thetax)
\end{align}
Finally, the RHO authors found that updating the conditional term to depend on $ x_{<i}$ was unstable, so they instead approximate this by a fixed model $ \thetadown$:
\begin{align}
     \approx \min_{x_1, \dots, x_n \subset \Dtrain} \sum_{i=1}^n  -\log \Pr(x_i| \thetadown) + \log  \Pr(x_i | \thetax)
\end{align}
\fi

Note that while both CoLoR-Filter and RHO-down approximate the posterior over parameters with a point estimate, RHO-down makes a few additional approximations. This is largely a result of RHO-down attempting to increase data diversity by using a sequential approach to selection that conditions on the previously selected data $ x_{<i}$. This is an understandable goal, but it introduces more approximations, can cause instability by creating a non-stationary data distribution, and is computationally expensive since the data selection is no longer parallelizable.
A continued discussion of the pros and cons of online selection is in \Cref{app:online}.

\paragraph{RHO-down + prior.} We also consider a version of the algorithm that we call ``RHO-down + prior'' that replaces $ \Ddown, \thetadown$ in the RHO-down algorithm with $ \Dprior \cup \Ddown , \thetapriordown$ to incorporate the prior information. This corresponds to conditioning on both $ \Dprior$ and $ \Ddown$ instead of only $ \Ddown$. Intuitively, this method can better leverage stronger features learned on the larger $ \Dprior$ to integrate the information from the small $ \Ddown$.

\section{Further Related Work}\label{sec:related}

We now discuss some related work, more broadly, with regards to active learning and data curation.

\textbf{Active \& Curriculum learning}. 
Our formulation of data selection has connections to classic and deep active learning \citep{houlsby2011bayesian, bickfordsmith2023prediction, kirsch2023a}, which are deeply rooted in optimal Bayesian experimental design \citep{lindley1956measure, rainforth2024modern}, whose goal is to select a set of experiments to optimize certain information criteria \citep{pukelsheim2006optimal} such as maximally reducing the uncertainty about model parameters. Various acquisition functions are proposed in deep learning regimes \citep{sener2018active, ash2019deep, ash2021gone} and most of them focus on label-efficient image classification.
Another line of recent techniques share deep methodological connections but emphasize the sub-selection of available data during training (rather than the collection of additional examples typically considered in active learning) and could thus be classified as curriculum learning \citep[e.g.][]{graves2017automated}. Among them, RHOLoss \citep{mindermann2022prioritized} seeks to select data based on the hold-out reference dataset from the same distribution as the training data. It has been later implemented in continual pre-training \citep{lin2024rho1} and vision domains \citep{evans2023bad, tack2024learning}. 
%Different from the above approaches, our work seeks to rigorously formulate the problem and derive tractable implementations, showing that raw pre-training data can be tailored to pre-train a good model for particular downstream tasks from scratch. 

\textbf{Data curation practices in pre-training}.
Though large-scale public web-crawled data are common data sources for pre-training models, low-quality, toxic, and uninformative content that can prevent successful pre-training is prevalent \citep{wenzek2020ccnet, elazar2023s, sorscher2022beyond, allenzhu2024physics}. 
Therefore, practitioners design sophisticated data pre-processing pipelines such as filtering \citep{brown2020language}, deduplication \citep{lee2022deduplicating}, and mixing \citep{touvron2023llama, touvron2023llama2} to improve the data quality.
Due to the immense scale, state-of-the-art pre-training datasets usually depend on simple heuristic filters \citep{raffel2020exploring, rae2021scaling, together2023redpajama} (e.g., URL, length, n-gram perplexity, fastest classifiers) that can be parallelized across CPU nodes.
Besides the above rule-based filtering, model-based filtering concerns using machine learning models to score and filter data, which has been proven to be effective in vision and vision-text domains \citep{schuhmann2022laion, abbas2023semdedup, fang2023data}.
Such approaches usually leverage a given trustworthy data source like Wikipedia or Books as the reference and contrast the raw data with it. 
Due to computational cost, models are often designed to be small such as n-gram \citep{xie2023data}, single-layer neural networks \citep{joulin2017bag, brown2020language}, k-means clustering \citep{tirumala2024d4}.
There is also a growing line of work illustrating that data quality is important in shaping model training from a variety of perspectives, such as increasing data scale \citep{hoffmann2022training, meta2023llama3} and using synthetic data \citep{gunasekar2023textbooks}.  
%Recent work DSDM leverages small-scale proxies and other techniques such as gradient compression \citep{engstrom2024dsdm}, minimax optimization \citep{xie2024doremi} to select data from corpora like C4, Pile.
%Our work focuses on exploring automated larger-scale model-based filtering - training small proxy/reference models for selecting influential data from post-filtered pre-training data C4 based on a particular validation set. 

% Finally, while outside the scope of this work, data curation techniques are commonly used in Continual Learning, albeit with the emphasis on reducing forgetting rather than improving downstream performance \citep[e.g.][]{rebuffi2017icarl, titsias2019functional, buzzega2020dark}.

% \textbf{Data quality in pre-training.}

%Our goal is not to refine existing data to improve their quality but rather to explore how to select a good data subset to achieve good performance on some target tasks of interest.

\section{Algorithms}\label{sec:algs}

\begin{algorithm}[h]
\caption{CoLoR-Filter}
\begin{algorithmic}[1]
\label{alg:color-filter}
  \REQUIRE Prior data $ \Dprior$, downstream data $ \Ddown$, training data $ \Dtrain$, budget $ n$,  subset size multiplier $ \tau$
  % \STATE Sample a subset $ \Dcandidate \subset \Dtrain$ of size $ \tau n$
  \STATE Pre-train $ \thetamarg$ on $ \Dprior$
  \STATE fine-tune to get $ \thetacond$ on $ \Ddown$ initialized from $ \thetamarg$
  \STATE Select a random subset $ D_\tau$ of size $ \tau n$ from $ \Dtrain$
  \STATE Select data:
  \vspace{-0.4cm}
  \begin{align*}
        S =  \texttt{bottom-}n_{x \in D_\tau}  -\log \Pr(x| \thetacond) + \log \Pr(x| \thetamarg)
    \end{align*}
    \vspace{-0.5cm}
  \RETURN Selected dataset $S$ to train $ \theta$ on.
\end{algorithmic}
\end{algorithm}

\subsection{From Derivations to Practical Algorithms}

In our experiments, we will consider four algorithms based on the above derivations. In this section we go through each of these in turn.

\paragraph{CoLoR-Filter.} Our proposed algorithm is presented formally in \Cref{alg:color-filter}. Compared to the derivation, the main difference is the introduction of $ \tau$, a hyperparameter that acts as a compute-performance trade-off controlling how expensive and aggressive the data selection is. Rather than selecting data from all of $ \Dtrain$, we take a random subset $ D_\tau$ of size $ \tau n$. Thus, larger $\tau $ subselect more aggressively, but at the cost of more computation. A full discussion of this cost is in \Cref{sec:compute}.

\paragraph{Conditional only.} As an ablation of CoLoR-Filter, we follow prior work \citep{evans2023bad} and include a baseline that only uses the conditional model to select data. Essentially, this is CoLoR-Filter if we always assume that $ \log \Pr(x|\thetamarg) = 0$ in Line 4 of \Cref{alg:color-filter}. 

\begin{algorithm}[h]
\caption{RHO-down}
\begin{algorithmic}[1]
\label{alg:rho}
  \REQUIRE Downstream data $ \Ddown$, train data $ \Dtrain$, budget $ n$, subset size multiplier $ \tau$, batch size $ b$
  \STATE Train $ \thetacond$ on $ \Ddown$
  \STATE Initialize a random $ \thetamarg_1$ and $ S = \varnothing$
  \FOR{$t \in [1, \dots, n/b]$}
  \STATE Randomly select a batch $ B_t \subset \Dtrain$ of size $ \tau b$
  \STATE Select data:
    \vspace{-0.4cm}
  \begin{align*}
        \bar B_t = \texttt{bottom-}b_{x \in B_t} - \log \Pr(x| \thetacond) + \log \Pr(x| \thetamarg_t)
    \end{align*}
    \vspace{-0.5cm}
    \STATE $S = S \cup \bar B_t$
    \STATE Update $ \thetamarg_t $ to $ \thetamarg_{t+1}$ by training on $ \bar B_t$
    \ENDFOR
  \RETURN Selected dataset $S$ to train $ \theta$ on.
\end{algorithmic}
\end{algorithm}

\paragraph{RHO-down.} We present a practical variant of RHO-down in \Cref{alg:rho} based on the derivation presented in \Cref{sec:derivations}. The main changes to make a practical algorithm are (1) the introduction of $ \tau$ as in CoLoR-Filter, and (2) performing the algorithm batch-wise instead of using single data points.

\paragraph{RHO-down + Prior.} We can also incorporate the prior data $ \Dprior$ into \Cref{alg:rho} by simply replacing Line 1 where $ \thetacond$ is trained on $ \Ddown$ with a procedure where we first pre-train $ \thetacond$ on $ \Dprior$ and then fine-tune it on $ \Ddown$.

\subsection{Computational Cost}\label{sec:compute}

To evaluate the computational cost of the various algorithms, we use units of ``model forwards'' per token where we assume that a backward pass is twice as expensive as a forward pass \citep{fleuret2023little}. Note that our 150m models take about 5e8 FLOPs per model forward of a single token \citep{hoffmann2022training, casson2023transformerflops}.
The cost of running the selection algorithms depends on $m, n, \tau$ and $ L$ defined as follows: 
$ m $ is the size of the prior data $ \Dprior$, $n $ is the size of the selected dataset $ S$, $ \tau$ is the hyperparameter controlling how aggressively we subselect data. 
Note that we assume that $ |\Ddown|$ is so small that the cost of training a model on $ \Ddown$ is negligible towards the total cost (and all the methods we consider just fine-tune a model once on $ \Ddown$). We will also be careful to note when computation can be done in parallel before training versus computation that must happen serially during a training run. Offline algorithms like CoLoR-Filter can take advantage of parallelism to improve efficiency. In this section, we go through each method in turn and aggregate the computational costs in \cref{tab:cost}.

\paragraph{Scale transfer.}
We also include another parameter $ L $ to cover the case where we select data using small models and use it to train a larger model \citep{evans2023bad}.
Specifically, $ L$ is the ratio of cost of one model forward of the \emph{large} target model compared to the small auxiliary models used for data selection.
For example, in our experiments, when we use 150 million parameter models to select data and then train a 1.2 billion parameter model on the resulting data, then $ L\approx 5.5$\footnote{Even though there are 8x as many parameters in the large model, the FLOP multiplier is less since the attention computations take the same number of FLOPs regardless of parameters.}. Training thus costs $ 3nL$ across all methods since we run a forward and backward for the large model on all $ n $ sequences.

\paragraph{CoLoR-Filter.} The cost of selection is $ 2\tau n$ forward passes. But, this selection process is \emph{entirely} parallelizable. Training the prior model costs $ 3m $ forwards since $ |\Dprior| = m$. And training a model on the selected data costs $ 3nL $ forward passes. So the total cost is $3m + 2 \tau n + 3nL$, but the $ 2 \tau n$ scoring computation can be done in parallel. 

\paragraph{Conditional Only.} The conditional-only method is almost the same as CoLoR-Filter, except we only need $ \tau n$ forward passes for selection since we only run one model over the data. The cost is thus $ 3m + \tau n + 3nL$, with $ \tau n$ being parallelizable.

\paragraph{RHO-down.} The cost of selection is still $ 2 \tau n$ forward passes. Then we need an additional $ 2n$ to backward the output model (since the forward is already handled during scoring). Note that we need to evaluate the marginal model online, so it is not parallelizable, but the conditional model is fixed and can be computed offline. So, the cost is $ 2 \tau n + 2n + 3nL$, and the $ \tau n$ conditional model computation can be done in parallel. 

\paragraph{RHO-down + Prior.} For the version with an added prior, we just add $ 3m$ cost for training the prior. Thus, the cost is $ 2\tau n + 2n + 3nL$ with $ \tau n$ parallelizable.

\begin{table}[t]
    \caption{Compute cost of the various algorithms measured in ``model forwards''. The total cost of selection and training on the selected data is the sum of all costs across a row. The variables are $ m = |\Dprior|$, $ n = |S|$, $ \tau$ is a hyperparameter that controls how aggressively we subselect, and $L $ is a multiplier of the cost of model forwards between the selection model(s) and the target model (approximately the ratio of parameter counts between the models). }
    \label{tab:cost}
    \centering
    \begin{tabular}{lllll}
    \toprule
        Method & Prior cost & Serial cost & Parallel cost & Training cost\\
        \midrule
        CoLoR-Filter &  $ 3m $ & $0$ & $2 \tau n$ & $3n L$ \\[3pt]
        Conditional Only & $ 3m $ & $ 0 $ & $\tau n$& $3n L$\\[3pt]
        RHO-down & 0 & $ \tau n + 2 n $ & $\tau n$& $3n L$\\[3pt]
        RHO-down + Prior & $ 3m $ &  $\tau n + 2n$ & $\tau n$& $3n L$\\[3pt]
        Random & 0 &  0 & 0 & $3n L$\\
    \bottomrule
    \end{tabular}
\end{table}

Overall, the methods all have comparable costs, with Conditional Only being the cheapest and RHO-down + Prior the most expensive. The main difference is that CoLoR-Filter and Conditional Only are easily parallelized while RHO-down and RHO-down + Prior are not. It should also be noted that when doing experimentation, offline methods like CoLoR-Filter also benefit from being able to re-use likelihoods multiple times, while RHO-based methods need to recompute the serial cost any time that some hyperparameter of the algorithm.

\section{Domain Transfer: a Simple Testbed}\label{sec:books}

\subsection{Setup}

\paragraph{Training.} We train language models with 150 million non-embedding parameters using the OLMo codebase \citep{groeneveld2024olmo} and following hyper-parameter choices from \citep{wortsman2024smallscale}. Unless otherwise noted, we use 150m models as the auxiliary models ($ \thetacond, \thetamarg$) as well as the target model $ \theta$. Full hyperparameters are described in detail in \Cref{sec:hyperparams}.

We take $ \Ddown$ to be a small dataset of 25 million tokens sampled from the Project Gutenberg Books data subset of Dolma \citep{soldaini2024dolma}, $ \Dprior$ to be a dataset of 3.1 billion tokens from C4 \citep{raffel2020exploring}, and $ \Dtrain$ to be all of C4. 
We select a dataset $ S $ of 3.1 billion tokens (which is approximately the ``chinchilla optimal'' amount for models of this size).
To get $ \thetapriordown$ or $ \thetadown$, we fine-tune or train for one epoch on $ \Ddown$. 

\paragraph{Evaluation.} To evaluate the efficacy of our data selection, we report cross-entropy loss of next token prediction on a held-out dataset $ \widetilde D_{\text{down}}$ from the same distribution as $ \Ddown$ (Books). 

\paragraph{Baselines.} The simplest baseline we consider is \textbf{Random} sampling, which has been shown to be a strong baseline for C4 pre-training \citep{engstrom2024dsdm}. We consider all four algorithms described in \Cref{sec:algs}: \textbf{CoLoR-Filter}, 
 \textbf{Conditional Only}, \textbf{RHO-down}, and \textbf{RHO-down + prior}. And as one extra baseline, we also include \textbf{DSIR} \citep{xie2023data} which estimates n-gram importance weights between $ \Dtrain$ and $ \Ddown$, and similarly has a parameter like $ \tau$ that controls how aggressively to subselect.

Note that while it is in a similar setting to ours, we do not include DSDM \citep{engstrom2024dsdm} as a baseline since there is no publicly available code and based on the appendix of that paper, it it much more computationally expensive than the methods we consider.

\subsection{Results}

\begin{wrapfigure}[13]{r}{0.5\textwidth}
    \centering
    \vspace{-1.5cm}
    \includegraphics[width=0.5\textwidth]{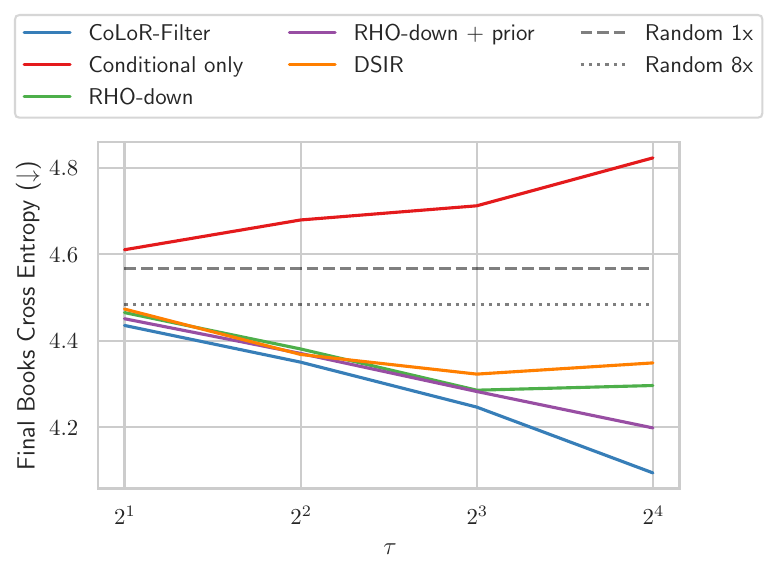}
    \vspace{-0.6cm}
    \caption{Scaling of final performance with $ \tau$ when targeting \textbf{Books} with 150m parameter models. %CoLoR-Filter scales best with $ \tau$.
    }
    \label{fig:books_tau}
\end{wrapfigure}

We first run the domain transfer experiments on 150m models, sweeping across $ \tau$ that controls the selected subset size. 
In \Cref{fig:books_tau} we plot how the final performance scales with $ \tau$ across methods. We see that CoLoR-Filter has the best scaling performance with increased $ \tau$, with no sign of saturation for $ \tau = 16$. We hypothesize that by using strong models to select the data, CoLoR-Filter is able to more effectively scale to larger $ \tau$ than the other methods.
In \Cref{fig:books} in \Cref{app:curves}, we plot the learning curves (evaluated on the held-out validation set) for the four methods introduced in \Cref{sec:algs}. There, we see especially clean scaling for CoLoR-Filter across the entire learning curve, substantially outperforming random selection with much less data, similar to \Cref{fig:1b}.

\begin{wrapfigure}[15]{r}{0.5\textwidth}
    \centering
    \vspace{-1.1cm}
    \includegraphics[width=0.5\textwidth]{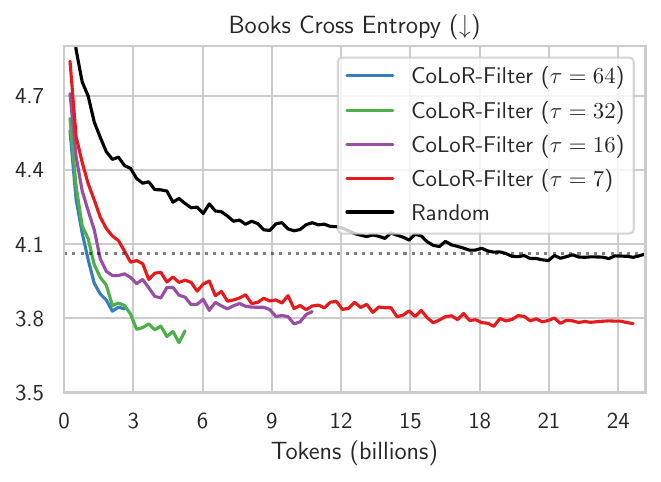}
    \vspace{-0.6cm}
    \caption{Scaling CoLoR-Filter with $ \tau$ when training 1.2b models with data selected by 150m models. Curves end when we exhaust the data in C4.}
    \label{fig:books_1b_tau}
\end{wrapfigure}

\paragraph{Scale generalization.} Finally, we also conduct an experiment in scale generalization (partially shown in \Cref{fig:1b}) using the data selected by our 150m auxiliary models to train a 1.2b target model. 
In \Cref{fig:books_1b_tau} we show learning curves for a sweep over $ \tau$.
We still see consistent gains as we scale $ \tau$ for a fixed number of training tokens. 
Interestingly, if we fix the total number of tokens we are \emph{selecting from} (i.e. where the lines end when we run out of C4), then the final performance with $\tau = 32$ is better than all other values of $ \tau$.
This shows how a strict subset of tokens can outperform a superset (e.g. $ \tau=16$).
We should also point out here the computational savings when using CoLoR-Filter.
As an example, consider $ \tau=16$ where we match the performance of 25 billion randomly selected tokens with about 1.5 billion filtered tokens. Considering the computational costs discussed above with $ L = 5.5$ and measuring $ n $ in billions of tokens, the total cost for training the CoLoR-Filter model is $ 3 m + 2\tau n + 3nL = 3 * 3.1 + 2 * 16 * 1.5 + 3 * 1.5 * 5.5 = 82 $ while the cost for training on 25 billion random tokens is $ 3NL = 3 * 25 * 5.5 = 412.5$, illustrating a more than 5x total compute savings to achieve the same performance on Books. A full plot visualizing the cost in FLOPs for all $ \tau$ is in \Cref{app:flops}.
%This only tells us the performance on one task, so in the next section we consider targeting a variety of tasks at once that are more substantially different from language modeling.

\section{Downstream Tasks}\label{sec:down}

\subsection{Setup}

\paragraph{Training.} We target the 8 tasks from the OLMo paper \citep{groeneveld2024olmo}: Hellaswag \citep{zellers2019hellaswag}, PIQA \citep{bisk2020piqa}, ARC-challenge and ARC-easy \citep{clark2018think}, Openbook QA \citep{mihaylov2018can}, SciQ \citep{welbl2017crowdsourcing}, BoolQ \citep{clark2019boolq}, and Winogrande \citep{sakaguchi2021winogrande}. Each of these datasets has a separate train split. We use these train splits to construct $ \Ddown$ as follows: for each question we concatenate the question and the correct answer formatted as a grammatical continuation. Overall, this results in a small $ \Ddown$ dataset of 7.4 million tokens. $ \Dprior$ and $ \Dtrain$ are the same as before. And we again get $ \thetapriordown$ by fine-tuning $ \thetaprior $ for one epoch on $ \Ddown$.

\begin{wrapfigure}[10]{r}{0.5\textwidth}
    \centering
    \vspace{-1.2cm}
    \includegraphics[width=0.5\textwidth]{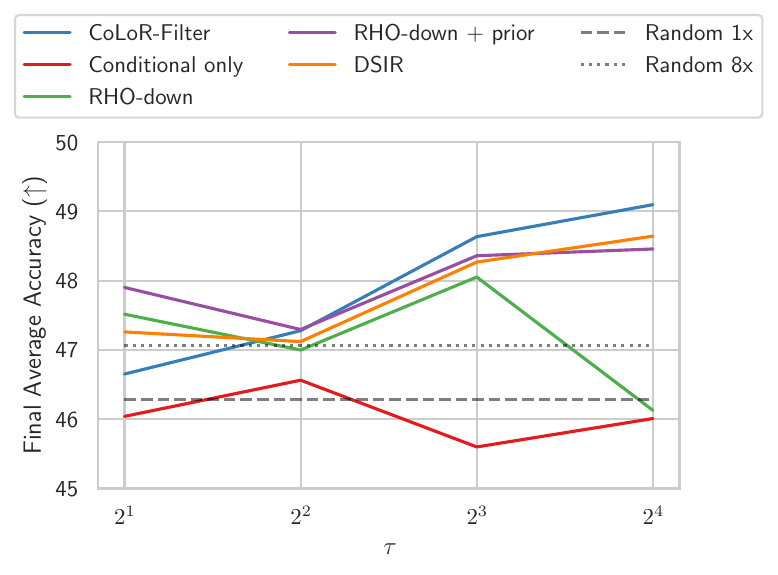}
    \vspace{-0.7cm}
    \caption{Final performance versus $ \tau$ on the suite of downstream tasks for 150m models. CoLoR-Filter scales the best with $ \tau$.}
    \label{fig:down_tau}
\end{wrapfigure}

\paragraph{Evaluation.} We evaluate on held-out data from each downstream task test or validation sets (using val if test is not publicly available). We use the evaluation procedure from OLMo \citep{groeneveld2024olmo} which follows \citep{eval-harness} for evaluating these multiple-choice tasks using the rank classification approach of \citet{brown2020language}. We report aggregat perfromance across tasks as well as the task-specific performance.

\paragraph{Baselines.} Same as in \Cref{sec:books}.

\subsection{Results}

\begin{figure}[t]
    \centering
    \includegraphics[width=0.85\textwidth]{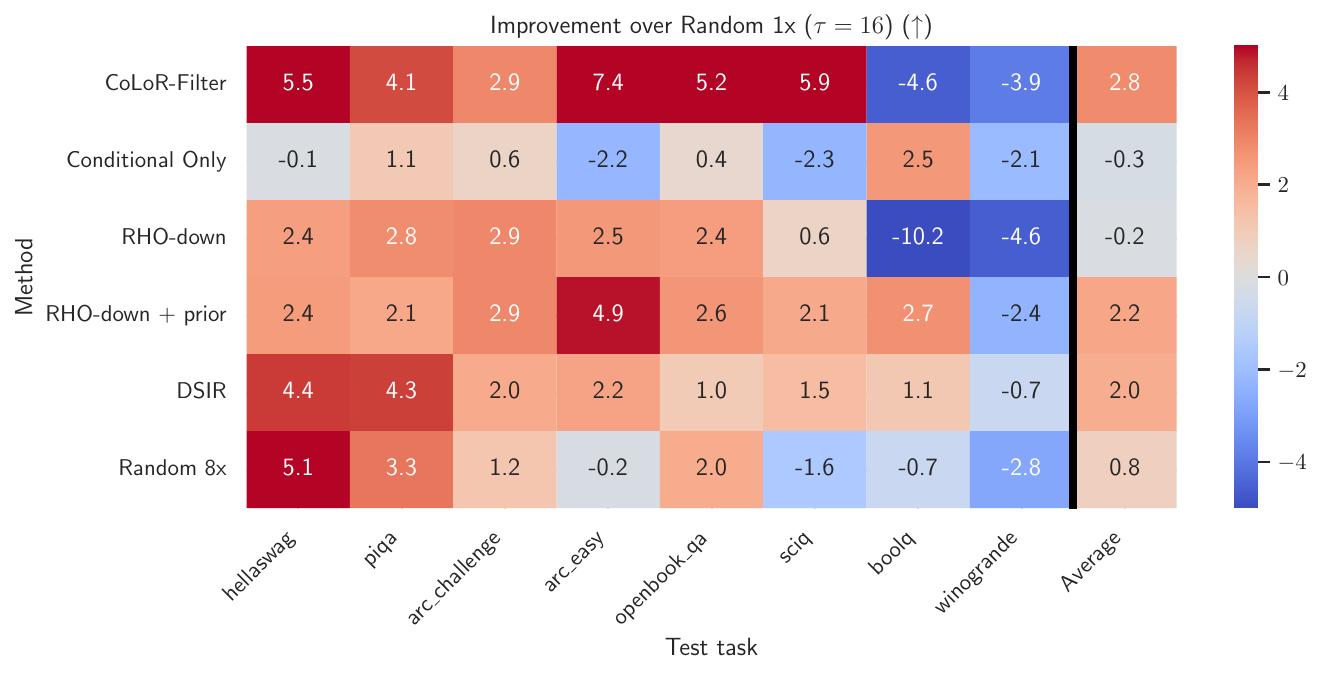}
    \vspace{-0.2cm}
    \caption{Performance improvement over training on an equivalent amount of random data broken down by task (except for Random 8x, which uses 8x more data). A table of results is in \Cref{app:table}.}
    \label{fig:heatmap}
\end{figure}

While the curves themselves are noisier now due to the noisier nature of accuracy evaluation on small datasets compared to cross entropy on a large one, the same trends hold as we saw for domain transfer to Books. CoLoR-Filter in particular is scaling the best as we increase $ \tau$. Other methods do not illustrate the same clean scaling as we increase $ \tau$, which is nearly linear on a log scale for CoLoR-Filter, as seen in \Cref{fig:down_tau}. Full learning curves are in \Cref{app:curves}.

We can also look at the performance broken down by task and illustrated relative to training on an equivalent amount (3.1 billion tokens) of randomly selected data for $ \tau = 16$ illustrated in \Cref{fig:heatmap}. We see especially large gains on Hellaswag, ARC easy, Openbook QA and SciQ and actually see performance decreases on BoolQ and Winogrande. However, we should note that at this scale and with all data selected from C4, we actually found BoolQ and Winogrande to be quite noisy and not even correlated with training on 8x as much random data, so it is not clear how much weight to place on those results. Across the other tasks, the gains of CoLoR-Filter over the baselines are clear. It is an interesting direction for future work to probe more deeply into how task-dependent the gains from targeted data selection can be.

\begin{wrapfigure}[17]{r}{0.5\textwidth}
    \centering
    \vspace{-0.6cm}
    \includegraphics[width=0.5\textwidth]{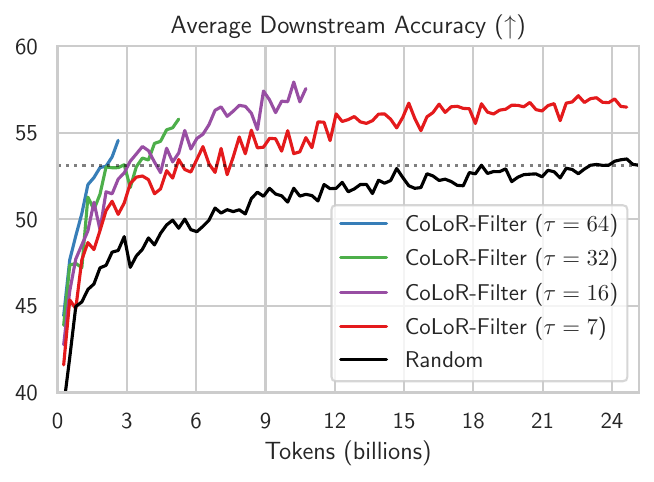}
    \vspace{-0.6cm}
    \caption{Scaling CoLoR-Filter with $ \tau$ when training 1.2b models with data selected using smaller 150m models. Curves end when we exhaust the data in C4.}
    \label{fig:down_1b_tau}
\end{wrapfigure}

\paragraph{Scale generalization.} We also consider scale generalization to a 1.2b target model and illustrate the full results of a sweep over $ \tau $ in \Cref{fig:down_1b_tau}. Again we find significant benefits of CoLoR-Filter across scales. A full table of per-task results is in \Cref{app:table}.  Again we notice that training on a strict subset of data can outperform a larger dataset.

We can again do out the calculation of computational savings for $ \tau = 16$. 
It now takes about 3 billion tokens for CoLoR-Filter to match the performance of training on 25 billion random tokens. This amounts to a total cost of $ 3 m + 2\tau n + 3nL = 3 * 3.1 + 2 * 16 * 3 + 3 * 3 * 5.5 = 154.8$, which is still an upwards of 2.5x reduction in compute to achieve the same average performance across the suite of tasks. A full plot visualizing the cost in FLOPs for all $ \tau$ is in \Cref{app:flops}.

\paragraph{Task generalization.} 
We can also test task generalization beyond the 8 tasks that were used to select the data on a few more tasks that test common sense reasoning \citep{wang2019superglue, socher-etal-2013-recursive, talmor2018commonsenseqa, sap2019socialiqa}. Results are presented in \Cref{tab:tasks} compared to a random model trained on 10x as much data. The performance indicates that the data selected by CoLoR-Filter are not overfit to the particular evaluation tasks, but captures some general notion of good data for a range of tasks.

\begin{table}[h]
    \caption{Task generalization for the 1.2b models with $ \tau = 64$.}
    \label{tab:tasks}
    \centering
    \resizebox{\textwidth}{!}{%
    \begin{tabular}{l|llllll}
    \toprule
Method & copa & rte & cb & sst2 & commonsense qa & social iqa \\ \midrule
Random (25b tokens) & \textbf{69.2} & 48.9 & 42.8 & 46.8 & \textbf{33.7} & \textbf{42.9}  \\
CoLoR-Filter ($\tau=64$, 2.5b tokens) & 65.8 & \textbf{52.6} & \textbf{46.0} & \textbf{55.8} & 32.6 & 42.7 \\
\bottomrule
    \end{tabular}
}
\end{table}

Note, we also conduct a few more experiments and ablations in the appendix: \Cref{app:id} considers using CoLoR-Filter in-distribution to target C4 loss, \Cref{app:batchwise} considers applying CoLoR-Filter batchwise rather than globally, \Cref{app:finetune} considers finetuning on $ \Ddown$ after targeted pre-training, \Cref{app:analysis} inspects some of the selected and excluded examples, and \Cref{app:fineweb} compared to FineWeb-edu \citep{penedo2024fineweb}.

\section{Discussion}

While fairly simple to derive and implement, we show that CoLoR-Filter is an effective method for data selection on C4, with promising scaling behavior up to 1.2 billion models. 
In our experiments, CoLoR-Filter continues to improve when only using 1 out of 64 data points considered for selection and generalizes from small auxiliary models to larger target models. This opens many potential lines of research. First, while we have considered targeted pre-training, it is possible that CoLoR-Filter could be extended to fine-tuning, continual pre-training, and more general open-domain pre-training. In particular, it is an interesting open question whether the lack of an explicit consideration of data diversity hinders CoLoR-Filter in any of these settings. Second, CoLoR-Filter could be applied to more challenging domains in language like code generation or even applied beyond the language domain to other modalities. Finally, there is plenty of work to be done to make the algorithm more efficient and to test the limits of scale generalization. 

\section*{Acknowledgments}

HZ is supported by an Eric and Susan Dunn Graduate Fellowship. SK acknowledges support from the Office of Naval Research under award N00014-22-1-2377 and the National Science Foundation Grant under award \#IIS 2229881. This work has been made possible in part by a gift from the Chan Zuckerberg Initiative Foundation to establish the Kempner Institute for the Study of Natural and Artificial Intelligence.

\bibliographystyle{plainnat}
\bibliography{references}

%%%%%%%%%%%%%%%%%%%%%%%%%%%%%%%%%%%%%%%%%%%%%%%%%%%%%%%%%%%%
\newpage
\appendix

% \section{Appendix / supplemental material}

% Optionally include supplemental material (complete proofs, additional experiments and plots) in appendix.
% All such materials \textbf{SHOULD be included in the main submission.}

\section{Learning curves for 150m models}\label{app:curves}
% Fig \ref{fig:down_lcs} and \ref{fig:books} denote the 

%\clearpage
\begin{figure}[h]
    \centering
    \includegraphics[width=\textwidth]{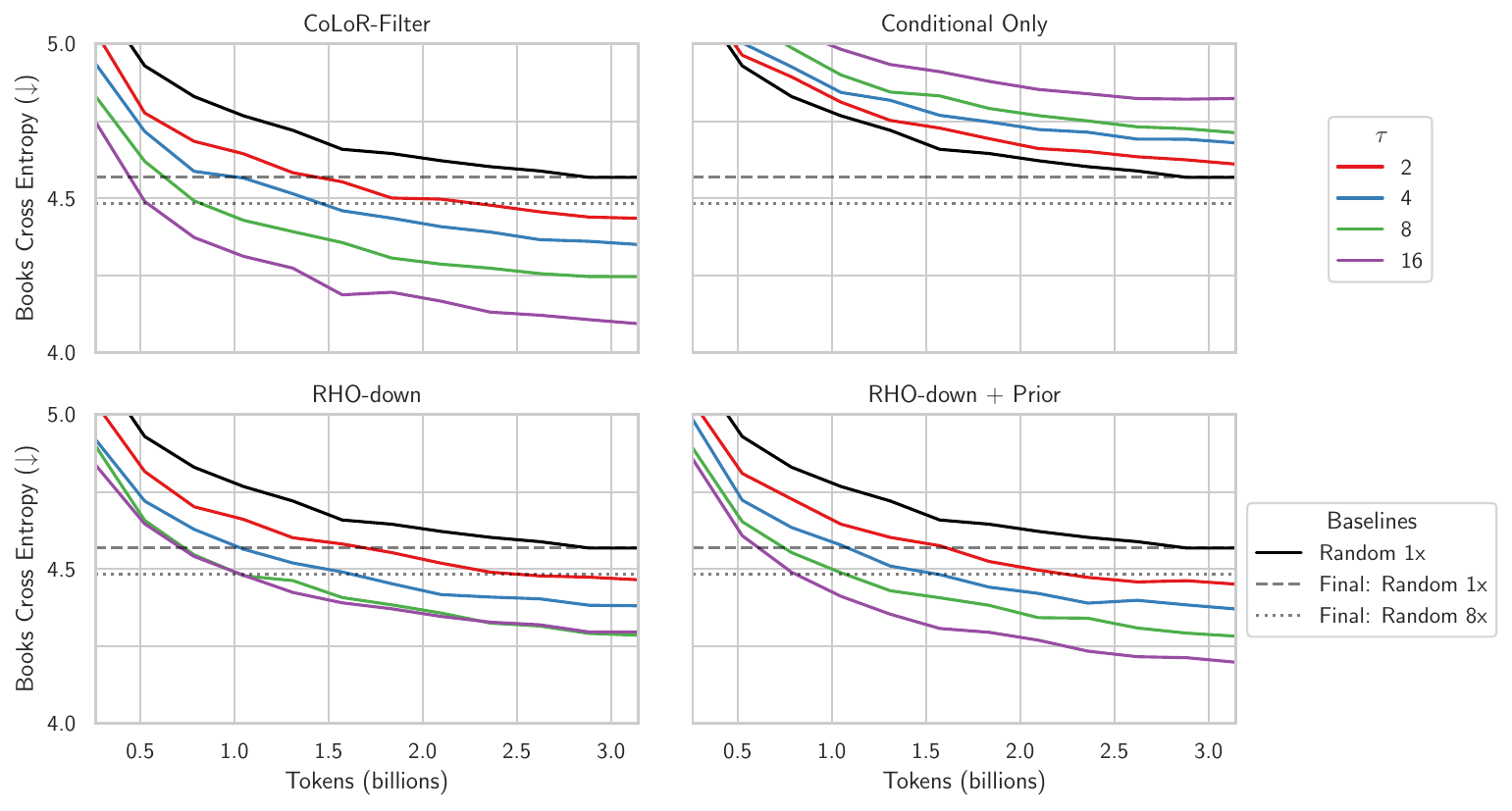}
    \caption{Sweeping over $ \tau$ when targeting \textbf{Books} from C4 for 150m models.}
    \label{fig:books}
\end{figure}

\begin{figure}[h]
    \centering
    \includegraphics[width=\textwidth]{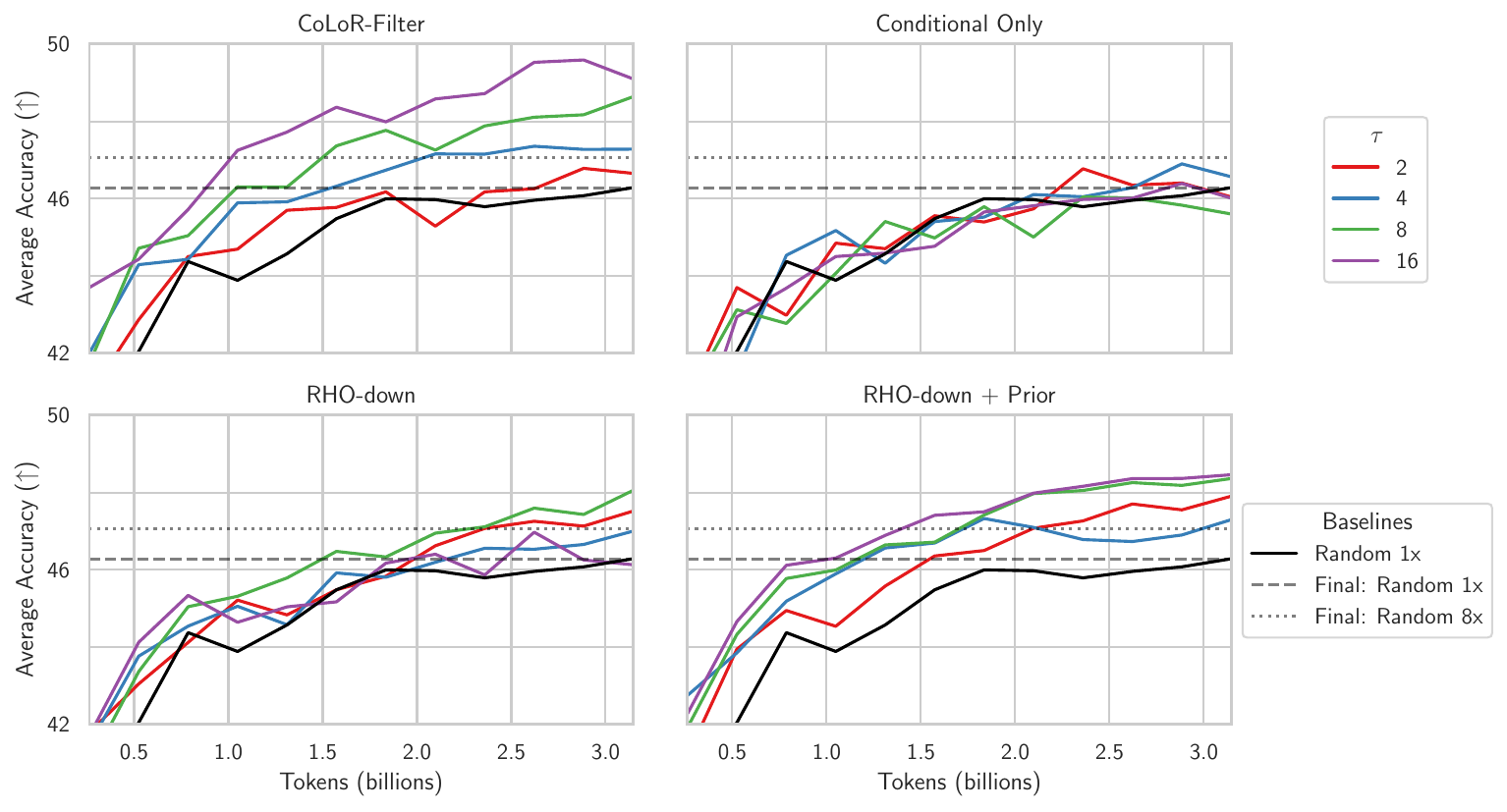}
    \caption{Sweeping over $ \tau$ and measuring average performance on all downstream tasks for 150m models.}
    \label{fig:down_lcs}
\end{figure}

\section{Tables of downstream results}\label{app:table}

\clearpage
\begin{table}[h]
    \caption{Performance for all tasks for 150m models for data selection with $ \tau = 16$.}
    \label{tab:150m-down}
    \centering
    \resizebox{\textwidth}{!}{%
    \begin{tabular}{l|llllllll|l}
    \toprule
Method & \makecell{hella-\\ swag} & piqa & arc-c & arc-e & \makecell{open-\\book qa} & sciq & boolq & \makecell{wino- \\ grande} & Avg \\ 
\midrule
Random 1x & 33.2 & 64.5 & 22.4 & 44.4 & 26.8 & 66.9 & 58.8 & \textbf{53.3} & 46.3 \\ 
CoLoR-Filter & \textbf{38.6} & 68.7 & \textbf{25.3} & \textbf{51.8} & \textbf{32.0} & \textbf{72.8} & 54.3 & 49.4 & \textbf{49.1} \\
Conditional Only & 33.0 & 65.6 & 23.0 & 42.2 & 27.2 & 64.6 & 61.4 & 51.1 & 46.0 \\
RHO-down & 35.5 & 67.3 & \textbf{25.3} & 46.9 & 29.2 & 67.5 & 48.6 & 48.7 & 46.1 \\
RHO-down + prior & 35.6 & 66.6 & \textbf{25.3} & 49.3 & 29.4 & 69.0 & \textbf{61.6} & 50.9 & 48.5 \\
DSIR & 37.6 & \textbf{68.8} & 24.4 & 46.6 & 27.8 & 68.4 & 59.9 & 52.6 & 48.3 \\
Random 8x & 38.2 & 67.8 & 23.5 & 44.2 & 28.8 & 65.3 & 58.1 & 50.5 & 47.1 \\
\bottomrule
    \end{tabular}
}
\end{table}

\begin{table}[h]
    \caption{Final performance for all tasks for 1.2b models. Note that the CoLoR-Filter models do not train on as many tokens since we exhaust all of the tokens in C4 with these settings of $ \tau$.}
    \label{tab:1b-down}
    \centering
    \resizebox{\textwidth}{!}{%
    \begin{tabular}{l|llllllll|l}
    \toprule
Method & \makecell{hella-\\ swag} & piqa & arc-c & arc-e & \makecell{open-\\book qa} & sciq & boolq & \makecell{wino- \\ grande} & Avg \\ 
\midrule
Random (25b tokens) & 52.9 & 73.0 & 26.1 & 53.7 & 32.8 & 75.5 & 56.7 & 54.3 & 53.1 \\
CoLoR-Filter ($\tau=7$, 25b tokens) & \textbf{62.3} & \textbf{75.6} & 29.7 & 60.3 & \textbf{38.0} & 79.7 & 48.3 & \textbf{58.0} & 56.5 \\
CoLoR-Filter ($\tau=16$, 10b tokens) & 59.3 & 75.4 & \textbf{31.7} & \textbf{62.7} & 36.2 & \textbf{81.0} & 57.7 & 56.4 & \textbf{57.6} \\
CoLoR-Filter ($\tau=32$, 5b tokens) & 54.8 & 74.3 & 29.4 & 60.9 & 35.4 & 78.4 & 59.1 & 54.1 & 55.8 \\
CoLoR-Filter ($\tau=64$, 2.5b tokens) & 49.3 & 73.2 & 28.9 & 59.7 & 35.6 & 77.1 & \textbf{59.8} & 53.0 & 54.6 \\
\bottomrule
    \end{tabular}
}
\end{table}

\section{Data diversity and online vs. offline selection}\label{app:online}

Much work on active learning focuses on ensuring that we select a diverse set of data points that cover the test distribution of interest. As explained in the main text, by making a point estimate of the parameters, CoLoR-Filter is simplifying the problem and sacrificing an explicit term for diversity in the objective. In practice, this seems to be saved by the facts that (1) C4 has already been deduplicated, (2) we still select a fairly large subset without replacement, and (3) an individual sequence contains diversity across tokens.

However, the fact that CoLoR-Filter sacrifices a notion of diversity in the objective is important to consider more deeply. 
Here, we derive what a loss-based algorithm for data selection that prioritizes diversity would look like and why it is computationally infeasible. 
Then we derive an approximation (that looks somewhat like RHOLoss \citep{mindermann2022prioritized}) and show how it is empirically unstable, as was also observed previously by \citep{mindermann2022prioritized}.

To derive a CoLoR-Filter-like algorithm that values diversity, we can start from \Cref{eq:bayes} by a greedy approximation where samples are added to $ S $ one (batch) at a time, like in RHO:
\begin{align}
     \approx \min_{x_1, \dots, x_n \subset \Dtrain} \sum_{i=1}^n  -\log \int_\theta\Pr(x_i| \theta)\Pr(\theta | \Ddown, x_{<i}) + \log \int_\theta \Pr(x_i | \theta)\Pr(\theta| x_{<i})
\end{align}
Note that this sort of greedy algorithm for subset selection has a long history in active learning \citep{das2018}, is actually theoretically sound in some cases \citep{Nemhauser1978AnAO}, and is used in prior work \citep{ash2021gone, mindermann2022prioritized}. Importantly, this algorithm still prioritizes selecting a diverse dataset. By conditioning on past data at step $ i$, the objective encourages the algorithm to select data that is different from data that has already been selected.

We can also make an empirical bayes version by adding $ \Dprior$:
\begin{align}
     \min_{x_1, \dots, x_n \subset \Dtrain} \sum_{i=1}^n  -&\log \int_\theta\Pr(x_i| \theta)\Pr(\theta | \Dprior, \Ddown, x_{<i}) \\ &+ \log \int_\theta \Pr(x_i | \theta)\Pr(\theta| \Dprior, x_{<i})
\end{align}

This is, of course, still intractable since it requires integrating the parameters. But, since we have already introduced the greedy algorithm that encourages diversity, if we now make the point estimate approximation, the incentive for data diversity remains. This results in:
\begin{align}\label{eq:online}
     \approx \min_{x_1, \dots, x_n \subset \Dtrain} \sum_{i=1}^n  -\log \Pr(x_i| \thetapriordownx) + \log  \Pr(x_i | \thetapriorx)
\end{align}

The thorny issue here is how to define $ \thetapriordownx$ and $ \thetapriorx$ in practice. In theory, these parameters should be trained on an iid sample from the union of the datasets. If we add the datapoints one at a time, the dynamics of the distribution shift over time can change how well the model corresponds to conditioning on the union of the dataset. But, this would require re-training the models every time we add a new $ x_i$ which is clearly impractical.

\begin{figure}
    \centering
    \includegraphics[height=0.31\textwidth]{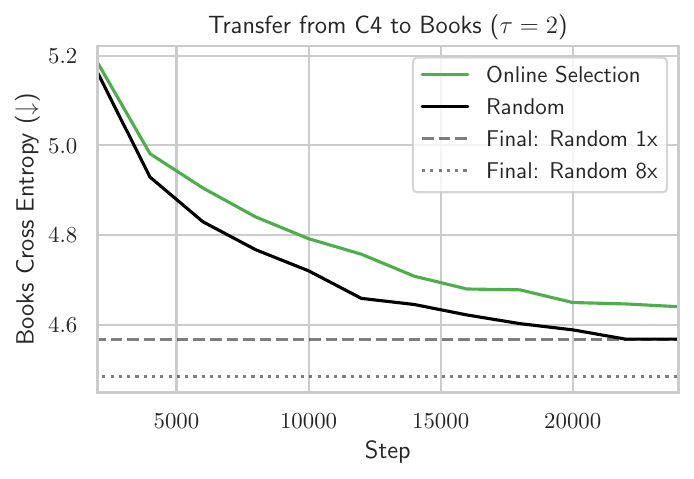}
    \hspace{1cm}
    \includegraphics[height=0.31\textwidth]{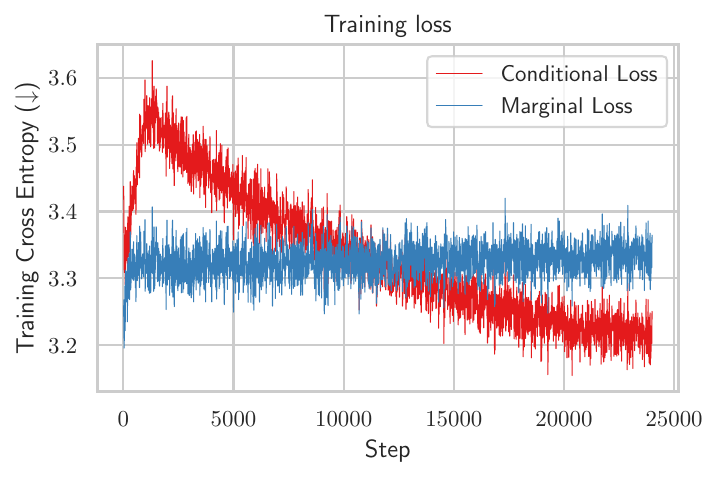}
    \caption{(Left) Performance of online selection with fine-tuning as outlined in \Cref{eq:online}. Online selection is worse than random. (Right) Training curves for the conditional and marginal models on the selected data $ S$. The conditional model faces training instability early on (associated with forgetting), and then eventually becomes better than the marginal on the selected data.}
    \label{fig:online}
\end{figure}

In practice, this encourages using a fine-tuning approach (as in RHO) where we continually fine-tune on the $ x_i$ as they are added. But when $ \Ddown$ is small and the data distribution changes over time, we can get catastrophic forgetting and unstable training dynamics. For these reasons, RHO avoids training the conditional model entirely (Appendix D of \citet{mindermann2022prioritized}). We also conduct an experiment on the Books task where we use this online fine-tuning algorithm that updates both the marginal and conditional models as we add data to $ S$. Results in \Cref{fig:online} show how the training is unstable and in fact performs worse than random.

Moreover, Note that the computational cost of even the cheapest fine-tuning algorithm is substantial compared to the algorithms in the paper. In particular, the serial cost is now $ 2 \tau n +  4n $ (as compared to $ \tau n + 2n$ for RHO) since we need to pass the full $ \tau n$ samples through both the conditional and marginal models. So this variant is clearly inferior in practice to the other approaches we consider.

\section{Compute cost for scale generalization}\label{app:flops}

\begin{figure}[h]
    \centering
    \includegraphics[width=0.8\textwidth]{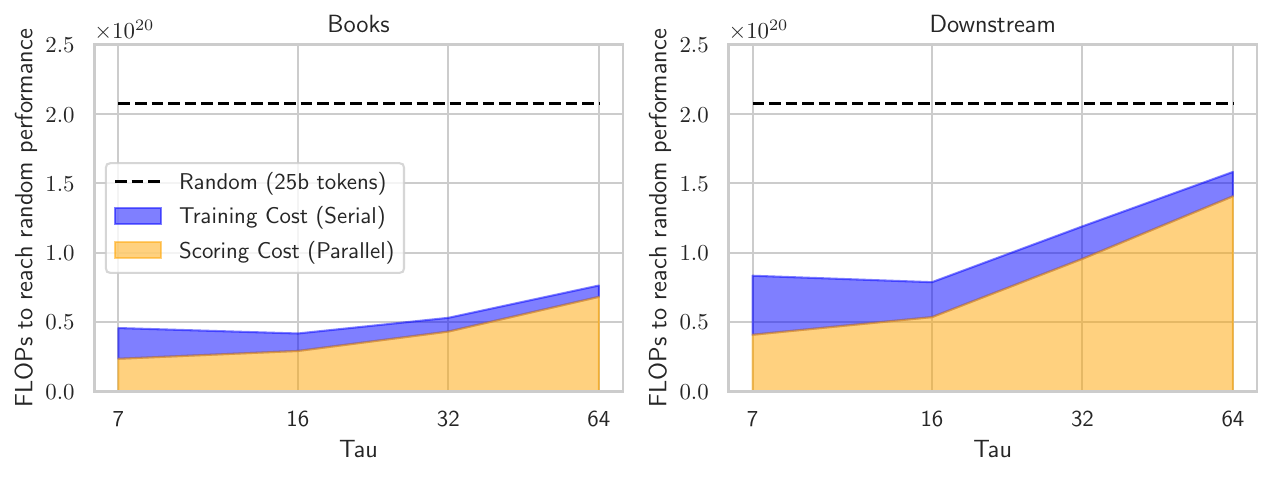}
    \caption{Costs in FLOPs to reach equivalent performance to the final random model trained on 25b tokens (i.e. cost until we reach the dotted line in \Cref{fig:1b}). We split cost into the scoring cost for filtering the data using the small auxiliary models and then training cost for the large model.}
    \label{fig:costs}
\end{figure}

In the main text we computed the cost for $ \tau = 16$ in terms of model forwards of 1 billion tokens. Here we can convert this to FLOPs and compute the cost for all values of $ \tau$. Results are in \Cref{fig:costs} showing the breakdown of costs into scoring FLOPs for running the small auxiliary models over the data and training FLOPs for training the large model. We measure the cost it takes to reach the final performance of the random model, i.e. until the CoLoR-filter learning curve crosses the dotted line in \Cref{fig:1b}. The main tradeoff is that lower $ \tau$ values require more scoring cost and less training cost because they are able to select better data.

We should also note that if multiple models are being trained with the same dataset, then this scoring cost can be amortized over those runs and the larger $ \tau$ values will look even better.

\section{Can we do data selection in distribution?}\label{app:id}

\begin{figure}[h]
    \centering
    \includegraphics[width=0.98\textwidth]{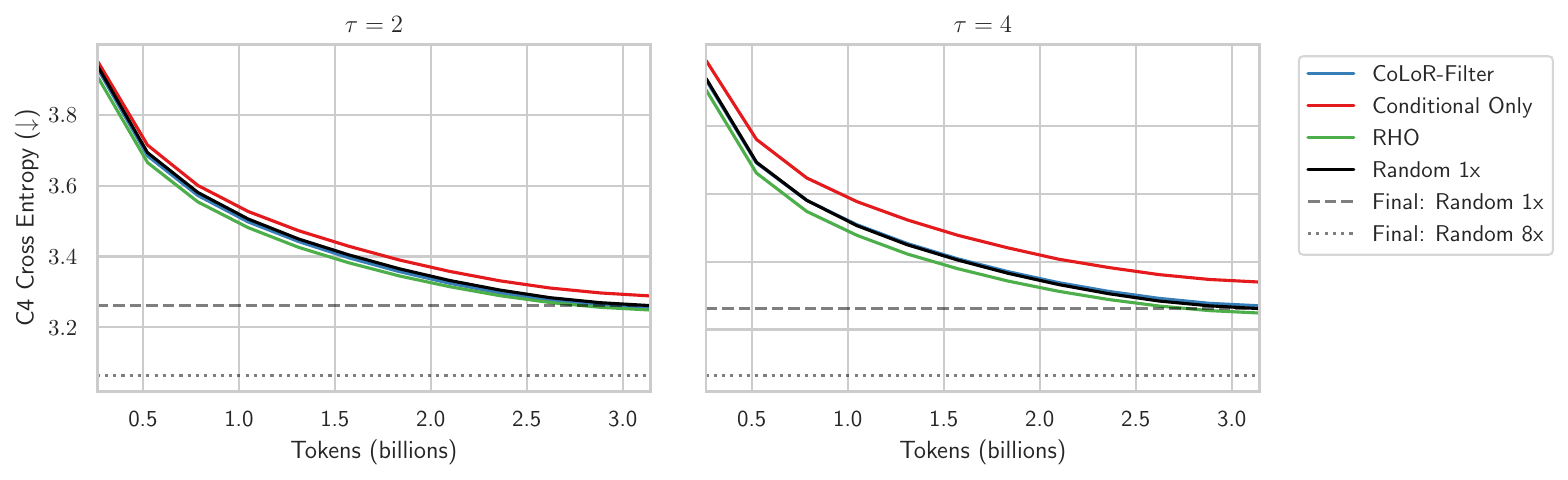}
    \caption{Using a sample of C4 as $ \Ddown$. RHO provides marginal gains here, while CoLoR-Filter does not provide gains at all. Conditional Only is worse than random. Scaling $ \tau$ does not change results as much as when we target downstream tasks.}
    \label{fig:in_dist}
\end{figure}

One obvious question raised by these data selection techniques is whether they can work in distribution, i.e. can we select data to make the iid loss on C4 go down faster? In \Cref{fig:in_dist} we present results for running this experiment with CoLoR-Filter as well as RHO and Conditional Only. Note that there is no difference between RHO and RHO + prior now (and we drop the ``down'' from the name) since the prior distribution and the downstream distribution are the same. To implement CoLoR-Filter in this setting, we just take two checkpoints from pre-training the prior model and call the earlier one (at 2.5b tokens) the marginal model and the later one (at 3.1b tokens) the conditional model. 

We find that in distribution selection does not work effectively with these methods. There are small gains to RHO loss, but here they are massively outweighed by the computational cost of the selection. CoLoR-Filter sees no gain at all over random and Conditional Only is worse than random. These preliminary results suggest why it is important to recognize that data selection (especially with these methods) will be most effective when we genuinely want to target a different distribution from $ \Dtrain$.

\begin{figure}[h]
    \centering
    \includegraphics[width=0.95\textwidth]{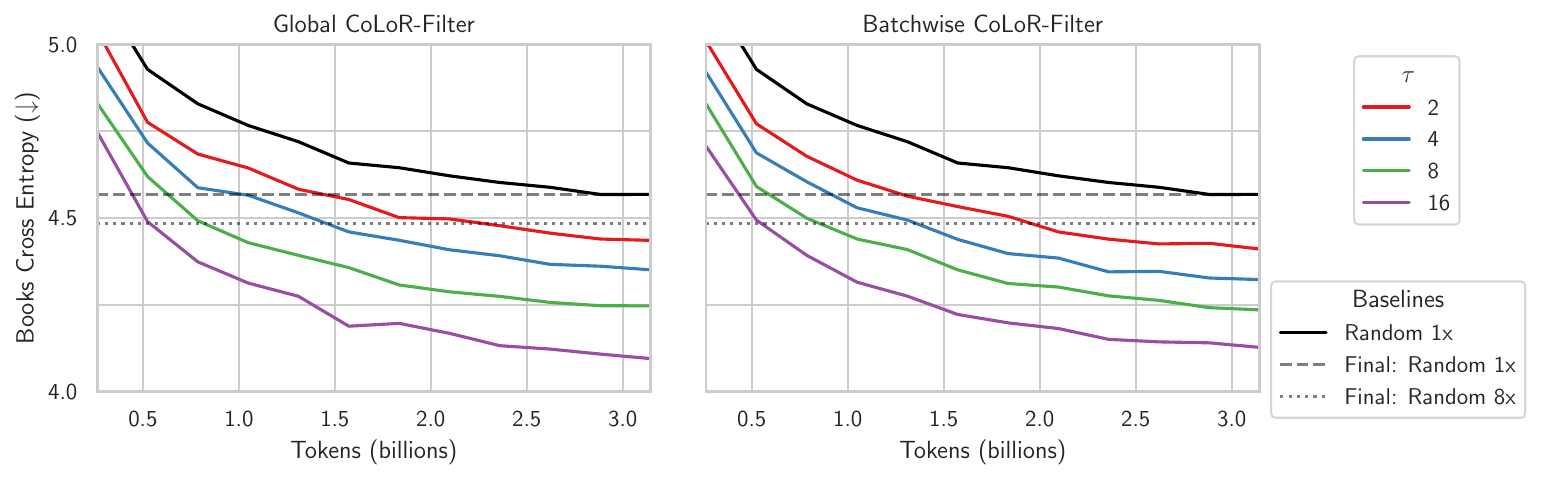}
    \caption{Comparison between global and batchwise variants of CoLoR-Filter on Books. The two perform nearly identically here.}
    \label{fig:batch}
\end{figure}

\section{Global vs. batchwise selection}\label{app:batchwise}

One more minor implementation aspect about CoLoR-Filter is that as presented in \Cref{alg:color-filter}, we do global selection where we take the best $ n$ data points across the entire train set, while in RHO-down in \Cref{alg:rho} selection is done batchwise. Here we ablate whether the ability to do global selection is actually helpful for CoLoR-Filter. Results in \Cref{fig:batch} suggest that there is not much difference between the two and at small $ \tau$, batchwise selection maybe even beat global selection. We provide this result to illustrate that CoLoR-Filter is fairly robust to how the selection is performed.

\section{Finetuning after targeted pre-training}\label{app:finetune}

One possible question about the targeted pre-training setting we consider is: what happens if we finetune on $ \Ddown$ after the targeted pre-training?

This is interesting since while the pre-trained models presented in the main text never have direct access to $ \Ddown$, the selection algorithm does. In this section, we also allow access to $ \Ddown$ after pre-training and then compare the final performance of the finetuned models that are pre-trained on random data vs. selected data.

First, in \Cref{tab:150m-books-fine} and \Cref{tab:150m-down-fine} we present finetuning results for the 150m models. We find that CoLoR-Filter data outperforms 8x as much random data after finetuning. Note that the conditional model that we use to guide the selection of CoLoR-Filter is equivalent to a model that has been pre-trained on 3B random tokens and then finetuned on the task. Thus, these results show that we are substantially outperforming the conditional model when both models are finetuned on the downstream data.

\begin{table}[h]
    \caption{Performance after finetuning on Books for different pre-trained 150m models. Note that the Random (3.1b tokens) model is equivalent to the conditional model used to select data with CoLoR-Filter ($\tau=16$).}
    \label{tab:150m-books-fine}
    \centering
    %\resizebox{\textwidth}{!}{%
    \begin{tabular}{l|c}
    \toprule
Pre-training data & Finetuned Books Val Cross Entropy \\ \midrule
Random (3.1b tokens) & 3.441\\
Random (25b tokens) & 3.357\\
CoLoR-Filter (3.1b tokens)  & \textbf{3.258}\\
\bottomrule
    \end{tabular}
\end{table}

\begin{table}[h]
    \caption{Held out performance after finetuning on downstream data for different pre-trained 150m models. Note that the Random (3.1b tokens) model is equivalent to the conditional model used to select data with CoLoR-Filter ($\tau=16$).}
    \label{tab:150m-down-fine}
    \centering
    \resizebox{\textwidth}{!}{%
    \begin{tabular}{l|llllllll|l}
    \toprule
Pre-training data & \makecell{hella-\\ swag} & piqa & arc-c & arc-e & \makecell{open-\\book qa} & sciq & boolq & \makecell{wino- \\ grande} & Avg \\ 
\midrule
Random (3.1b tokens) & 34.4 & 66.6 & 24.8 & 51.7 & 28.0 & 89.9 & \textbf{65.6} & \textbf{53.1} & 51.8 \\
Random (25b tokens) & \textbf{39.5} & 69.8 & \textbf{29.2} & 53.9 & 30.2 & \textbf{91.4} & 64.2 & 52.9 & 53.9 \\
CoLoR-Filter (3.1b tokens) & 39.2 & \textbf{71.1} & 29.1 & \textbf{55.3} & \textbf{33.2} & 90.0 & 65.1 & 51.6 & \textbf{54.3} \\
\bottomrule
    \end{tabular}
}
\end{table}

\begin{table}[h]
    \caption{Performance after finetuning on Books for different pre-trained 1.2b models. Note that the conditional model that selects data is only 150m parameters.}
    \label{tab:1.2b-books-fine}
    \centering
    %\resizebox{\textwidth}{!}{%
    \begin{tabular}{l|c}
    \toprule
Pre-training data & Finetuned Books Val Cross Entropy \\ \midrule
Random (25b tokens) & 3.074\\
CoLoR-Filter (2.6b tokens) & \textbf{2.964}\\
\bottomrule
    \end{tabular}
\end{table}

\begin{table}[h]
    \caption{Held out performance after finetuning on downstream data for different pre-trained 1.2b models.}
    \label{tab:1.2b-down-fine}
    \centering
    \resizebox{\textwidth}{!}{%
    \begin{tabular}{l|llllllll|l}
    \toprule
Pre-training data & \makecell{hella-\\ swag} & piqa & arc-c & arc-e & \makecell{open-\\book qa} & sciq & boolq & \makecell{wino- \\ grande} & Avg \\ 
\midrule
Random (25b tokens) & \textbf{55.3} & 74.6 & 35.2 & 63.0 & 35.8 & \textbf{94.6} & \textbf{72.0} & \textbf{62.5} & \textbf{61.6} \\
CoLoR-Filter (2.6b tokens) & 53.4 & \textbf{76.1} & \textbf{35.8} & \textbf{65.6} & \textbf{36.8} & 93.2 & 66.6 & 58.9 & 60.8 \\
\bottomrule
    \end{tabular}
}
\end{table}

Next, we present results for the 1.2b models in \Cref{tab:1.2b-books-fine} and \Cref{tab:1.2b-down-fine}. We find that the CoLoR-Filter model outperforms or is competitive with training on about 10x as much data randomly selected data. We should also note that the CoLoR-Filter models are now dramatically outperforming the 150m conditional models that were used to filter the data, showing positive scale transfer of data selection.

\section{Hyperparameters}\label{sec:hyperparams}

\begin{table}[h]
        \caption{150m model parameters, based on \cite{wortsman2024smallscale, groeneveld2024olmo}}
        \label{tab:150m-hyper}
        \centering
    \begin{tabular}{l|l}
    \toprule
        Parameter & Value \\
        \midrule
         Residual dimension & 1024\\
         Depth & 12 \\
         MLP hidden dimension & 4096\\
         Activation & GeLU\\
         Head dimension & 64\\
         Context length & 512\\
         Positional encoding & RoPE\\
         Biases & False\\
         Normalization & PyTorch Layernorm\\
         QK normalization & True\\
         Precision & Mixed, bfloat16\\
         Tokenizer & GPTNeox\\
         \bottomrule
    \end{tabular}
\end{table}

\begin{table}[h]
    \caption{1.2b model, based on \cite{wortsman2024smallscale, groeneveld2024olmo}. Only reporting differences from 150m.}
    \label{tab:1.2b-hyper}
    \centering
    \begin{tabular}{l|l}
    \toprule
        Parameter & Value \\
        \midrule
         Residual dimension & 2048\\
         Depth & 24 \\
         MLP hidden dimension & 8192\\
         \bottomrule
    \end{tabular}
\end{table}

\begin{table}[h]
    \caption{Training parameters, based on \cite{wortsman2024smallscale, groeneveld2024olmo}}
    \label{tab:train-hyper}
    \centering
    \begin{tabular}{l|l}
    \toprule
        Parameter & Value \\
        \midrule
        Optimizer & Adam\\
        Batch size & 256\\
        Learning rate & 1e-3\\
        Schedule & Linear warmup, cosine decay\\
        Warmup steps & 5\% of total steps\\
        z-loss coefficient & 1e-4\\
        Weight decay & 0.0\\
        $ \beta_1$ & 0.9\\
        $ \beta_2$ & 0.95\\
        $ \epsilon$ & 1e-15\\
    \bottomrule
    \end{tabular}
\end{table}

\newpage

\section{Inspecting the selected data}\label{app:analysis}

In this section, we conduct some basic analysis of the data that is selected by CoLoR-Filter. We leave a full analysis to future work, but here we provide some high level statistics about the distributions of the scores of the conditional vs. marginal models and some representative examples from the datasets.

\subsection{Distribution of scores}

First, we simply plot the CDFs of the conditional loss reduction (CoLoR) score function used to select the data. We find that there are relatively few outliers and the CoLoR scores are fairly concentrated and normally distributed. Moreover, we note that the mean CoLoR in both experiments is positive, meaning that the conditional model actually has higher losses on the datapoints in C4 than the marginal model. This makes sense because the conditional model has been finetuned on $ \Ddown$ which is out of distribution relative to C4.

\begin{figure}[h]
    \centering
    \includegraphics[height=0.31\textwidth]{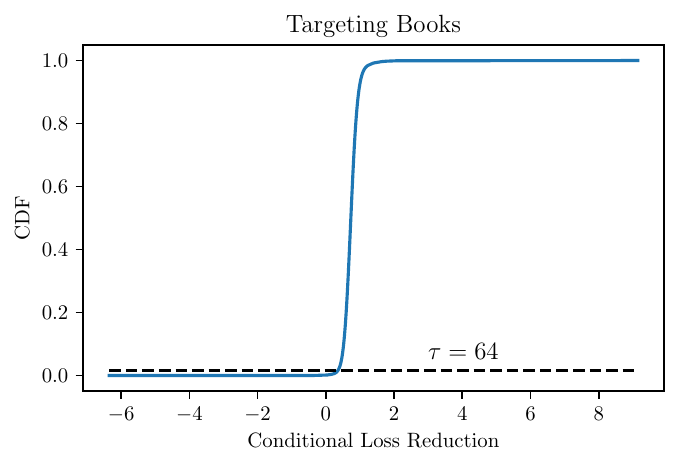}
    \hspace{1cm}
    \includegraphics[height=0.31\textwidth]{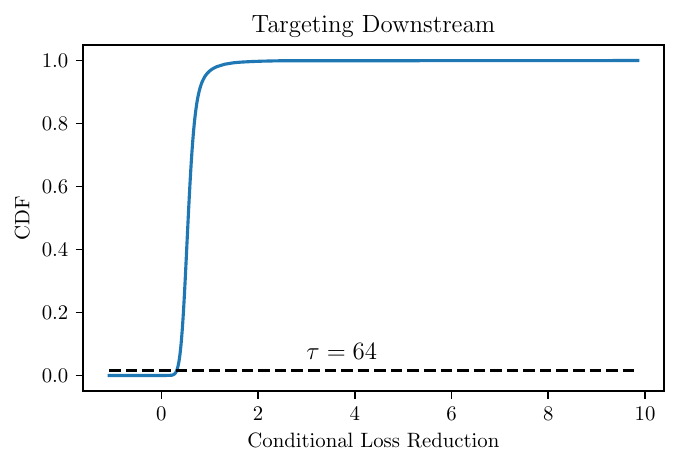}
    \caption{CDFs for the conditional loss reduction (CoLoR), i.e. $ -\log \Pr(x|\thetapriordown) - (-\log \Pr(x|\thetaprior))$. The dashed line highlights the cutoff point for $ \tau = 64$. We select the points with the lowest CoLoR.}
    \label{fig:cdfs}
\end{figure}

\subsection{Representative examples}

Now we just list a few representative examples to give a flavor for the types of outliers that exist under our ranking of sequences and the sorts of typical sequences that are selected versus excluded. The sequences are sampled randomly from different quantiles of the distribution and we shorten all the sequences so that they fit more easily on the page. 

\Cref{fig:books_outliers} shows outliers when targeting Books and \Cref{fig:books_typical} shows more typical examples when targeting Books. Generally, we found that the documents with very high scores contain things like old English, poetry, and tables of contents that are particularly unusual in books compared to the rest of the internet. Other things like fiction and dialogue are also highly scored. Negative outliers typically have things like poorly encoded text or advertisements.

\Cref{fig:down_outliers} shows outliers when targeting downstream tasks and \Cref{fig:down_typical} shows more typical examples when targeting downstream tasks. Here the patterns are less clear since the target tasks are more diverse, but we did observe many scientific and wiki-style documents with high scores as well as some descriptions of physical interactions that may be useful for common sense tasks. Again, the negative outliers tend to have things like poorly encoded text or advertisements. 

\begin{figure}[h]
    \begin{subfigure}[b]{0.45\textwidth}
        \centering
        {\small
        \texttt{
AS now shall ye wyt, what tyme of the day ye shall angle. From the begynning of Maye vntill it be September: the byting tyme is early in the morow from four of the clocke vnto eyght of the clocke, at after none from foure to eyght also, but not so good as in the mornyng, and if it be a colde wynde and a lowryng day, it is muche better than a cleere daye. Also many poole fysshes will byte best in the morne tyde.
And if ye se in any tyme of the day the Troute or greylyng lepe angle to him with a dub according to the same moneth. And where the water ebbeth and floweth: the fish wyll byte in some place at the ebbe and in some place at the flud after they haue restyng}}
        \caption{Good outlier, CoLoR = -0.35}
    \end{subfigure}
    \hfill
        \begin{subfigure}[b]{0.45\textwidth}
        \centering
        {\small
\texttt{
??????????????????????????????? ???????????????????????????????? ?????????????????????????????????
??????????????????????????????????
??????????????????????????????????
???????????????????????????????? ???????????????????????????????????
????????????????????????????????????
?????????????????????????????????????
?????????????????????????????????????
?????????????????????????????????????
?????????????????????????????????????
???????????? ????????????????????????? m88 ???????????????????????? ???? m88 ??????????????????????????????????????
??????????????????????????? ???? ?????????????????????????????????? ??????????????????????????????? ? ???????????????????
}}
        \caption{Bad outlier, CoLoR = 5.45}
    \end{subfigure}
    \caption{Examples of outliers when targeting \textbf{Books}. Examples are sampled randomly from the top or bottom 1000 sequences. The positive outlier is written in an older dialect of English which may be related to some documents in the Project Gutenberg corpus, while the negative outlier appears to be poorly encoded.}
    \label{fig:books_outliers}
\end{figure}

\begin{figure}[h]
    \begin{subfigure}[b]{0.45\textwidth}
        \centering
        {\small
        \texttt{
    C: Mrs Mackenzie, was there ever a time when you felt like you could just hop on a plane and make that flight down to the next State to be with your boys?
B: Oh my dear, yes. I feel sometimes as if I’m twenty and so fit and active and I can do whatever I want to do and then I remember, good grief, I’m 86, you old fool, you can’t do that. I wish I could just fly down there and live with them all together just how it was when they were little and I was their Mum and they followed me because I was so bright and cheery and smart and active and all the things that I’m not now. Oh, I’m so sorry, listen to me. Maybe I’m just losing my marbles, what do you think, dear?
C: Smiling – Imagine if I waved a magic wand and miraculously you were twenty again. What would you see yourself doing Beryl. Is it ok if I call you Beryl?
}}
        \caption{Sequence from best 3\%, CoLoR = 0.40}
    \end{subfigure}
    \hfill
    \begin{subfigure}[b]{0.45\textwidth}
        \centering
        {\small
    \texttt{Chamber of Commerce and other business venues, such as the Gwinnett Civic \& Convention Centers and is an ideal working environment for commercial businesses and corporations in Northeast Atlanta. The prominent location is on a heavily wooded, landscaped 6.5 acre site fronting on I-85. The exterior features green-tinted thermal glass and the entrance features a curtain wall glass leading into a granite-floored lobby with vaulted ceilings. Gwinnett County is home to leading Fortune 500 companies, drawn by its reputation as a commerce and technology hub, providing businesses with a regional market of five million people.                     SERVPRO of Gurnee can simplify the restoration process by handling both the initial water damage mitigation and rebuilding the affected areas. Having one qualified company for the entire process can save time and keep costs low.
    }}
    \caption{Sequence from median 3\%, CoLoR = 0.73}
    \end{subfigure}
    \caption{Examples of more typical documents when targeting \textbf{Books}. First a document from the top 3\% that would be selected with $ \tau = 32$, and then a document that scores near the median of all documents. The selected document is fictional dialogue while the median document is an advertisement.}
    \label{fig:books_typical}
\end{figure}

\begin{figure}[h]
    \begin{subfigure}[b]{0.45\textwidth}
        \centering
        {\small
        \texttt{
 among the pinacoderm are the ostia that allow entry of water into the body of the sponge. These pores have given the sponges their phylum name Porifera—pore-bearers. In some sponges, ostia are formed by porocytes, single tube-shaped cells that act as valves to regulate the flow of water into the spongocoel. In other sponges, ostia are formed by folds in the body wall of the sponge. Between the outer layer and the feeding chambers of the sponge is a jelly-like substance called the mesohyl, which contains collagenous fibers. Various cell types reside within the mesohyl, including amoebocytes, the “stem cells” of sponges, and sclerocytes, which produce skeletal materials. The gel-like consistency of mesohyl acts like an endoskeleton and maintains the tubular morphology of sponges.
The feeding chambers inside the sponge are lined by choanocytes (“collar cells”).}}
        \caption{Good outlier, CoLoR = -0.46}
    \end{subfigure}
    \hfill
        \begin{subfigure}[b]{0.45\textwidth}
        \centering
        {\small
\texttt{
 *** **********.
****** *** ***, *** ******* **** **** ** ******** ******* plates ** ****** ** ** **-** *** (*** ******* ** tested), ***** ******* ********.
*** ** *** ***, *** ********* *.* ********* ******* ***** capture ****** ******** ******** ****** ** **** ** **** ******, >10 ***, *** ******, **+ ***, **** ** ****** ***** or ****, *** ** ***** ****** *** *** **** **** field ** ****, ***** **'.
***** ******* ********, *** ******** ** ***** ****** ****** ****** to ******* ****** ** ****** **** ** **** ****** ** night, ****** ******* ******* *** ********.
******** ******** ** */****, *** ******** ****** ******** ******** **** front *** **** ****** ****** ** *** *** **** ******. However, **** ******* ******* *** ********** ** *** ***** ** night, ****** ** **** ****** *** ******* ************ ** *** scene.
}}
        \caption{Bad outlier, CoLoR = 5.36}
    \end{subfigure}
    \caption{Examples of outliers when targeting \textbf{downstream} tasks. Examples are sampled randomly from the top or bottom 1000 sequences. The positive outlier is a scientific document that could be relevant for tasks like SciQ, while the negative outlier appears to be poorly encoded.}
    \label{fig:down_outliers}
\end{figure}

\begin{figure}[h]
    \begin{subfigure}[b]{0.45\textwidth}
        \centering
        {\small
        \texttt{
summer plans. After thinking for a while I decided to spend my summer in Squamish, where I would work for the Admissions Team. However, due to a very large number of students interested to work on campus and a limited number of work positions, I ended up not getting a job on campus. I was very upset indeed and I began to think that there were not any job openings elsewhere, which would then result in me travelling back home.
Surprisingly, there were many job opportunities in the Squamish community. Since Quest University Canada hosted a job fair on campus I, along with all the students, had the chance to meet local businesses that were looking for summer employees. It was a great opportunity to network and give my resume to the ones that interested me. }}
        \caption{Sequence from best 3\%, CoLoR = 0.33}
    \end{subfigure}
    \hfill
        \begin{subfigure}[b]{0.45\textwidth}
        \centering
        {\small
\texttt{
Can I install PDF Stacks on more than one computer?
The license key is valid for only one device and is non-transferable. You can obtain additional license key(s) by placing an order.
How do I use PDF Stacks?
Click "File" and then "Import Folder"
Once you import the PDF files, your files will be copied into PDF Stacks for easier ability to read, search, organize, take notes, print and share. Any questions, ask us!
How do I create collections (virtual binders) and match/tag my documents for better organization?
It's easy. Watch the video for creating collections and tagging documents.
Can multiple users access the same documents or can I access and sync my documents through multiple devices?
}}
        \caption{Sequence from median 3\%, CoLoR = 0.55}
    \end{subfigure}
    \caption{Examples of more typical documents when targeting \textbf{downstream} tasks. First a document from the top 3\% that would be selected with $ \tau = 32$, and then a document that scores near the median of all documents. The selected document appears to be a journal entry while the median document is software documentation}
    \label{fig:down_typical}
\end{figure}

\clearpage

\section{Comparison to Fineweb-Edu}\label{app:fineweb}

Concurrent to our initial work, \citet{penedo2024fineweb} released FineWeb-edu, a classifier for educational content that can filter the FineWeb dataset. Here we provide a comparison between CoLoR-Filter and this classifier-based approach.

\begin{figure}[h]
    \centering
    \includegraphics[width=0.3\linewidth]{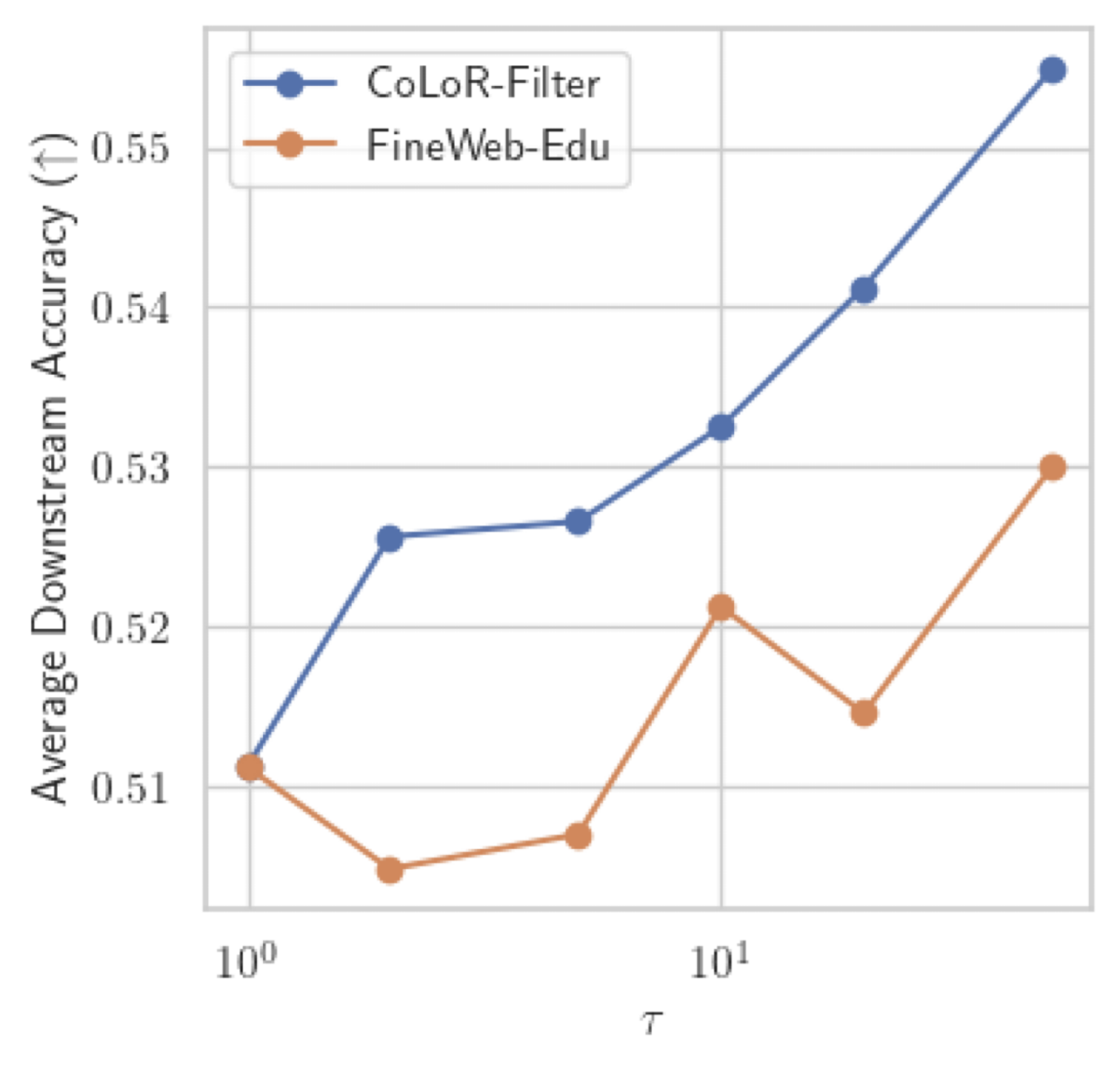}
    \caption{A comparison of the performance of 680m models trained on 10B tokens selected with various $\tau$ between CoLoR-Filter and FineWeb-edu.}
    \label{fig:fineweb}
\end{figure}

Specifically, we re-implement the CoLoR-Filter pipeline on top of the Fineweb dataset and with slightly smaller auxiliary models (125m) to make a more fair comparison to FineWeb-edu. Then we compare on the same suite of 8 downstream tasks over various settings of $ \tau$ using the two scores: CoLoR-Filter or the FineWeb-edu classifier. We then train larger models (680M parameters) for 10B tokens of selected data. Results are shown in \cref{fig:fineweb}. We find that CoLoR-Filter consistently outperforms FineWeb-edu, which is not so surprising since we are doing more targeted data selection by specifically targeting the downstream NLP tasks rather than a general notion of ``educational content''.

\section{Broader Impact}\label{sec:broader}
The development of the CoLoR-Filter for data selection has notable broader impacts on both machine learning and society. It enhances efficiency in language model training, leading to reduced computational resources and environmental footprint, while its scalability democratizes access to high-performing models. The method's success in diverse downstream tasks promises advancements in fields like medical text processing and legal analysis. However, it also raises concerns about dataset bias, necessitating continuous evaluation and updates. Future research should focus on ensuring models do not inherit biases from the selected training data, extending applications, improving efficiency, and implementing safeguards to maximize societal benefits while minimizing risks.

\section{Compute resources}\label{sec:cluster}
All training is conducted on an internal cluster using H100 GPUs. On one GPU, each 150m training run for 3.1b tokens takes about 4 hours, running the auxiliary models offline and in parallel can be faster. Training the 1.2b model to completion takes about 2 days on 4 GPUs.

\end{document}